\documentclass{article}

\PassOptionsToPackage{numbers, compress}{natbib}
\PassOptionsToPackage{hyphens}{url}
\usepackage[main, final]{neurips_2026}

\usepackage[utf8]{inputenc} 
\usepackage[T1]{fontenc}    
\usepackage{CJKutf8}        
\usepackage{hyperref}       
\usepackage{url}            
\usepackage{booktabs}       
\usepackage{amsfonts}       
\usepackage{nicefrac}       
\usepackage{microtype}      
\usepackage[table]{xcolor}  
\usepackage{enumitem}
\usepackage{amsmath, amssymb, amsthm}
\usepackage{algorithm, algorithmic}
\usepackage{wrapfig}
\usepackage{multirow, makecell}
\usepackage{graphicx}
\usepackage{subcaption}
\usepackage{fvextra}
\usepackage{tcolorbox}
\tcbuselibrary{breakable, skins}
\usepackage{minitoc}

\mtcsetfont{parttoc}{section}{\small\bfseries}
\mtcsetfont{parttoc}{subsection}{\small}

\definecolor{roleOrch}{HTML}{3A5A91}    
\definecolor{roleExec}{HTML}{C44B6F}    
\definecolor{roleRecall}{HTML}{2C7A39}  
\definecolor{roleJudge}{HTML}{6A4C93}   
\definecolor{tcBody}{HTML}{3D3D3D}      
\definecolor{tcSlot}{HTML}{B06A31}      

\fvset{fontsize=\scriptsize, breaklines=false, breakautoindent=false}
\DefineVerbatimEnvironment{vbPrompt}{Verbatim}%
  {formatcom={\color{tcBody}},commandchars=\^^A\^^B\^^C}

\newtcolorbox{promptbox}[2][]{%
  breakable, enhanced,%
  colback={#2!5}, colbacktitle=#2, coltitle=white, colframe=#2,%
  fonttitle=\bfseries\sffamily\small, title=#1,%
  boxrule=0.5pt, arc=1.5pt,%
  toptitle=2.5pt, bottomtitle=2.5pt,%
  left=5pt, right=5pt, top=3pt, bottom=3pt,%
  before upper={\parindent=0pt},%
  overlay first={%
    \draw[#2,line width=0.5pt] (frame.south west) -- (frame.south east);},%
  overlay middle={%
    \draw[#2,line width=0.5pt] (frame.north west) -- (frame.north east)
                               (frame.south west) -- (frame.south east);},%
  overlay last={%
    \draw[#2,line width=0.5pt] (frame.north west) -- (frame.north east);}}

\hypersetup{colorlinks=true, citecolor=blue!50!black, linkcolor=blue!50!black, urlcolor=blue!50!black}

\newcommand{\obs}[2]{\textit{\textbf{\underline{Observation #1:} #2}}}

\title{DAGent: Evaluate-then-Grow Planning \\ for Deep Research Agents}

\author{%
  Hanwen Liu$^{1,2}$ \qquad Yuanfu Sun$^{1,2}$ \qquad Qiaoyu Tan$^{2}$\thanks{Corresponding author.} \\[3pt]
  $^{1}$New York University \\
  $^{2}$New York University Shanghai \\[3pt]
  \texttt{\{hl6825, ys6310, qt2097\}@nyu.edu}
}

\begin{document}

\doparttoc          
\faketableofcontents

\maketitle

\begin{abstract}
Deep research tasks require agents to navigate large knowledge spaces, synthesize evidence across many sources, and adapt their plans as intermediate findings emerge. Directed acyclic graph (DAG)-based multi-agent systems are well suited to this setting because they support parallel execution and isolate each sub-task within a focused dependency context. Yet existing DAG-based agents typically instantiate a task-level plan before execution and then repair the graph only after failures or missing evidence are observed. This \textit{Plan-then-Patch} strategy is brittle for deep research: the system commits most strongly when its evidence is weakest, and later revisions often waste computation on branches that should not have been planned in the first place. To address this brittleness, we propose \textbf{DAGent}, a DAG-based multi-agent framework that introduces \textbf{Evaluate-then-Grow} incremental planning. Instead of committing to a full DAG upfront, an Orchestrator grows the task graph one batch at a time, conditioning each new expansion on confidence and uncertainty signals from completed nodes. To support long-horizon evidence use without overloading each sub-task, DAGent maintains a hierarchical context layer that propagates compact QueryDocs by default while preserving full execution traces for on-demand recall. The cleanly recorded DAG topology in turn admits structural RL signals that an outcome-only recipe cannot define; we instantiate this with \textbf{DAGRPO}, a GRPO adaptation that injects topology-conditioned credit on Executor rollouts and a structural compliance regularization on Orchestrator plans. Across BrowseComp-Plus, GAIA, and xbench-DeepSearch, DAGent surpasses the strongest open-source baseline by 5.3 / 5.8 / 2.0 points at the Qwen3-235B-A22B scale, with the lead replicating across four open-source backbones from four vendors and extending to GPT-5 at 327K context. At the Qwen3-8B scale, DAGRPO further improves over a same-budget outcome-only GRPO baseline by 3.0 average Pass@1 points. A same-architecture comparison further shows that incremental, evidence-conditioned planning reaches higher accuracy at lower per-task token, tool-call, and step footprints than its Plan-then-Patch counterpart, so the gains come from more targeted evidence expansion rather than additional computation. Code is available at \mbox{\url{https://github.com/hanwenliu6825/DAGent}}.

\end{abstract}

\section{Introduction}
\label{sec:intro}

The rapid advancement of large language models has given rise to a new class of information-intensive tasks broadly referred to as \emph{deep research}: tasks that require an AI system to autonomously navigate large knowledge spaces, synthesize information across dozens or hundreds of sources, and resolve complex queries through extensive multi-step reasoning. A growing body of work has established that \emph{DAG-based multi-agent systems} are an effective architectural foundation for such tasks, decomposing a query into a directed acyclic graph of sub-tasks so that independent branches execute in parallel and each agent operates within a focused context window informed only by its dependencies rather than the full upstream history. The paradigm has been applied to parallel function calling~\citep{kim2024llmcompiler}, web search~\citep{chen2025mindsearch}, multi-hop retrieval~\citep{verma2025planrag,chen2025logicrag} and long-horizon task decomposition~\citep{li2026tdp}.

Deep research, however, places greater demands on the \emph{planning mechanism} than traditional information search: intermediate findings can redirect subsequent investigation and invalidate assumptions that looked reasonable at the outset, making static or semi-static planning a poor fit. Recent DAG-based deep research systems acknowledge this through execution-time editing. Flash-Searcher \citep{qin2026flashsearcher} produces its complete plan in a single decomposition call and then periodically updates it, and FlowSearch \citep{hu2025flowsearch} initializes its knowledge-flow graph over several planner iterations, all of which precede execution, and then lets a separate refiner rewrite the graph through six edit operations. In both systems a plan that covers the whole task is committed before any node has executed, and later edits patch that standing plan. We term this paradigm \textbf{Plan-then-Patch}: it makes its largest planning commitment at precisely the moment when the system's understanding of the problem is shallowest, and the resulting trajectories mix plan commitment with plan revision, degrading answer accuracy and wasting compute as task complexity grows.

\textit{\textbf{Therefore, we argue that benefiting fully from DAG-based decomposition in deep research requires both the planning paradigm and the reinforcement learning (RL) objective to reflect the fact that the DAG is built incrementally}}.

\begin{wrapfigure}{r}{0.5\linewidth}
\centering
\includegraphics[width=\linewidth]{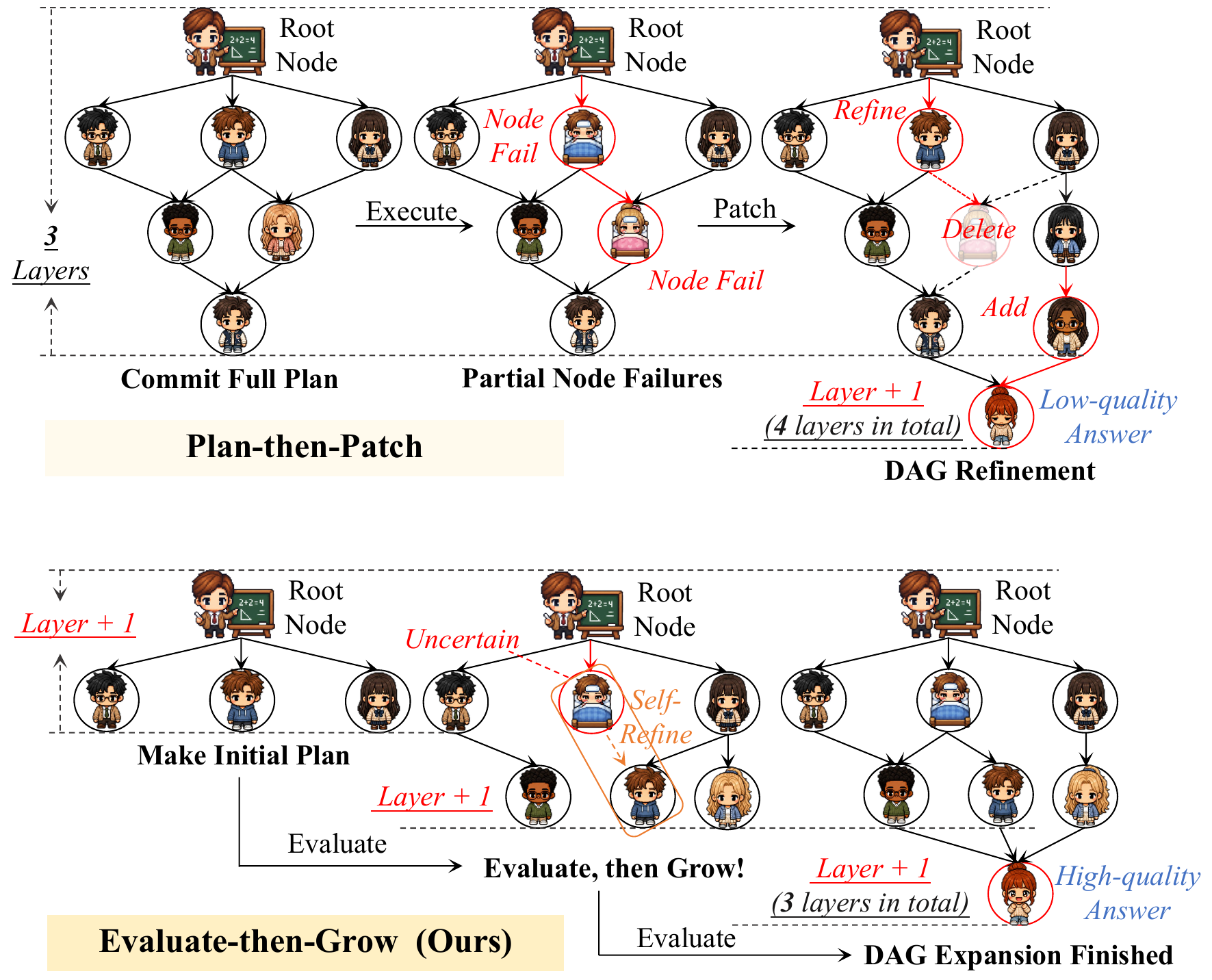}
\caption{Comparison of Plan-then-Patch (top) and Evaluate-then-Grow (bottom) paradigms.}
\label{fig:paradigm}
\end{wrapfigure}

To this end, we propose \textbf{DAGent}, a DAG-based multi-agent framework that replaces Plan-then-Patch with \textbf{Evaluate-then-Grow} incremental planning; Figure~\ref{fig:paradigm} contrasts the two paradigms. Instead of committing to a full task decomposition upfront, an Orchestrator grows the task DAG one batch of nodes at a time, using feedback from each completed node to decide what to plan next; each newly added node is then executed in parallel by a ReAct Executor that runs a standard ReAct loop \citep{yao2023react} equipped with domain-specific tools. To support long-horizon evidence use without overloading each sub-task, DAGent further maintains a hierarchical context layer on top of this incremental DAG: each node exposes a compact summary for default propagation to downstream nodes, and a complete execution trace that a downstream ReAct Executor can retrieve on demand. Figure~\ref{fig:dagent} illustrates the overall architecture.

The cleanly recorded DAG topology that Evaluate-then-Grow produces enables reinforcement learning signals that outcome-only recipes cannot define: rollouts at different DAG positions contribute unequally to the final answer, yet an outcome-only GRPO recipe assigns the trajectory-level reward uniformly across all rollouts. We exploit this with \textbf{DAGRPO}, a DAG-conditioned RL adaptation that augments a GRPO-style training recipe~\citep{shao2024deepseekmath} with two structural signals: a topology-conditioned credit on Executor rollouts based on whether their outputs feed the final synthesis, and a structural compliance regularization on Orchestrator plans that fail validation. Both signals are uniquely determined by the recorded trajectory because the DAG is append-only: no node is ever deleted or rewired, so the context a node saw when it ran coincides with its ancestry in the final graph.

We evaluate DAGent on BrowseComp-Plus, GAIA, and xbench-DeepSearch, covering local retrieval, web search, and Chinese-language deep research. Across backbones from Qwen3-8B to Qwen3-235B-A22B, DAGent consistently outperforms the strongest training-free baseline on all three benchmarks, with the authors' released implementation of FlowSearch included as a controlled baseline at Qwen3-32B; the lead replicates across four open-source backbones from four vendors, and the framework remains competitive with state-of-the-art agent systems when scaled to GPT-5 at 327K context. RL training with DAGRPO further improves Qwen3-8B DAGent over its role-separated GRPO counterpart at the same compute budget. Our major \textbf{contributions} are summarized below.
\begin{enumerate}[noitemsep,leftmargin=*]
\item \textbf{Evaluate-then-Grow planning.} We introduce an incremental DAG planning paradigm that departs from Plan-then-Patch: the task DAG is grown on the fly through structured node state evaluation rather than committed to upfront and reactively edited. To the best of our knowledge, DAGent is the first deep research agent to condition each planning decision on structured per-node feedback that summarizes the sub-task outcome, rationale, and reliability.

\item \textbf{DAGRPO.} We present a DAG-conditioned reinforcement learning variant for deep research agents that exploits the incremental DAG through topology-conditioned credit assignment and a structural compliance regularization on Orchestrator plans.

\item \textbf{Broad empirical advantage.} We show that the advantage of DAGent is broad rather than concentrated: it holds across BrowseComp-Plus, GAIA, and xbench-DeepSearch, replicates across four open-source backbones from four vendors, and persists when scaled to GPT-5 with a 327K context. We further show that the structural signals of DAGRPO add gains on top of an outcome-only GRPO baseline at the same compute budget, with off-chain credit sensitivity confirming that the gain is not a hyperparameter artifact.
\end{enumerate}

\section{Related Work}
\label{sec:related}

\subsection{Deep Research Agents}
\label{sec:rw_agents}

Long-horizon deep research stresses both the context window and the planning depth of an LLM, motivating two complementary lines of work that extend the linear reasoning-and-acting paradigm \citep{yao2023react}. The first manages context across a long linear trajectory: summary-based methods periodically compress the history into a summary state \citep{hu2025hiagent, wu2025resum, kang2025acon, lu2025supo}, and folding-style methods maintain context through branch-and-fold operations over linear sessions \citep{sun2025foldagent, ye2025agentfold, shao2025foldact}. The second decomposes the task into a structured plan: static DAG planners parallelize tool calls or multi-hop retrieval \citep{kim2024llmcompiler, chen2025mindsearch, verma2025planrag, chen2025logicrag, parallelsearch2025}, and Plan-then-Patch deep research agents commit a task-covering plan before any node executes, in one decomposition call \citep{qin2026flashsearcher} or over several planner iterations \citep{hu2025flowsearch}, and then edit it during execution; concurrent variants explore per-node localized DAGs \citep{li2026tdp} and hierarchical-outline planning \citep{li2025webweaver}. A growing set of proprietary deep research products \citep{openai2025deepresearch, monica2025manus, skywork2025deepresearch, moonshot2025kimi, metaso2025deepresearch} and open-source frameworks \citep{qiu2025alita, zhu2025oagents, pang2025browsemaster} report results on public deep research benchmarks. Across these lines, planning either operates on linear trajectories with no multi-parent dependencies, or commits the structured plan before any node has executed. DAGent differs by growing the DAG one batch at a time under structured per-node signals: the confidence and uncertainty fields of each completed node drive every new expansion of the graph.

\subsection{Structure-Aware RL-Based Optimization for LLM Agents}
\label{sec:rw_rl}

Group-based policy gradient methods such as GRPO \citep{shao2024deepseekmath} and DAPO \citep{yu2025dapo} have become the dominant online RL backbone for LLM agents, with adaptations to summary-based and fold-based long-horizon agents \citep{wu2025resum, sun2025foldagent}. Applied to multi-turn agents, these outcome-only recipes assign the trajectory-level reward uniformly across rollout tokens and ignore intra-trajectory structure. A parallel line of work injects finer-grained structural signals into the advantage at different granularities: turn-level \citep{wei2025multiturn}, step-level via anchor-state grouping \citep{feng2025gigpo}, tree-structured rollouts \citep{ji2025treegrpo}, and tool-use DAGs with graph-based rewards or dependency-aware grouping \citep{lu2025orchdag, wu2025gap}; for deep research planning specifically, DeepPlanner \citep{fan2025deepplanner} shapes GRPO advantages with an entropy-based term without exploiting DAG topology. Concurrent work on multi-agent RL assigns credit by role \citep{hong2025mgrpo}, by counterfactual removal of an agent \citep{li2026ccpo}, or over a learned communication topology among a fixed set of agents \citep{cang2026graphgrpo}; Appendix~\ref{appendix:concurrent_rl} contrasts these with DAGRPO. A topology-conditioned credit over the answer-feeding chain is uniquely determined by the recorded trajectory only when the graph is append-only. Plan-then-Patch trajectories do not preserve this property, since a node can be planned, executed, and later deleted or rewired, so its ancestry in the final graph need not match what the synthesis actually read. DAGRPO instantiates this signal on top of Evaluate-then-Grow.

\section{Method}
\label{sec:method}
\begin{figure}[ht]
  \centering
  \includegraphics[width=\linewidth]{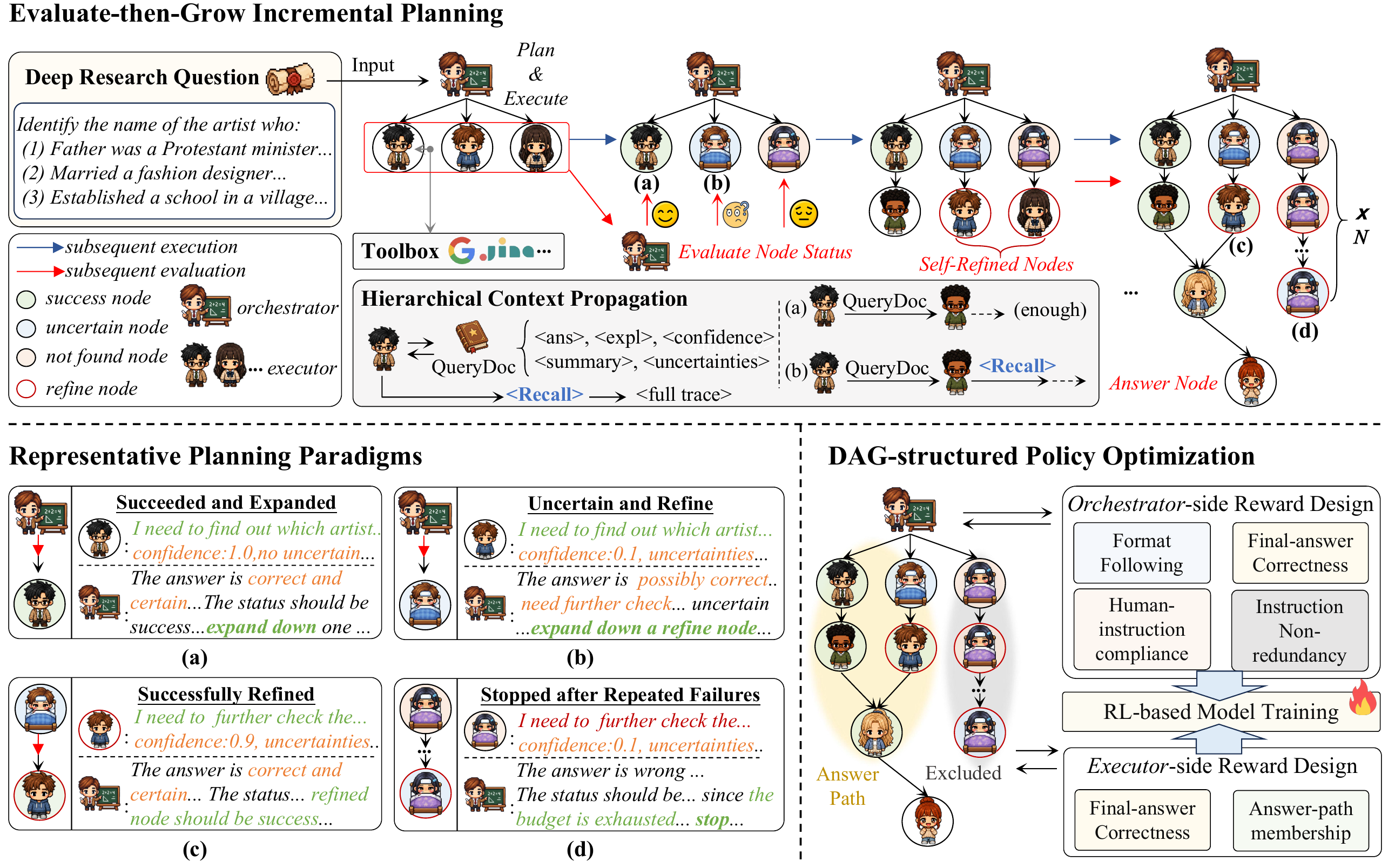}
  \caption{Overview of DAGent. The top panel shows Evaluate-then-Grow planning, where the Orchestrator evaluates each executed layer and expands the DAG accordingly. Nodes marked as Uncertain or Not Found trigger refine nodes for further verification, while child nodes receive the required context from their parents until an answer node is produced. The bottom-left panel shows four representative planning cases (a--d), marked at the corresponding nodes of the top panel: a node that succeeded and was expanded, a node that was uncertain and refined, a successfully refined node, and a node stopped after its refine budget was exhausted. The bottom-right panel summarizes DAG-structured policy optimization, where Orchestrator and Executor rewards are computed separately and optimized jointly, as detailed in \S\ref{sec:dagrpo}.}
  \label{fig:dagent}
\end{figure}

Figure~\ref{fig:dagent} illustrates the DAGent architecture. Section~\ref{sec:dagent} introduces Evaluate-then-Grow incremental planning together with the hierarchical context layer it supports; Section~\ref{sec:dagrpo} introduces DAGRPO, a DAG-conditioned RL adaptation that exploits the topology \S\ref{sec:dagent} records turn by turn.

\subsection{Evaluate-then-Grow Incremental Planning}
\label{sec:dagent}
Deep research planning is a sequential decision problem under evolving evidence: each completed sub-task may redirect the investigation or invalidate prior assumptions. Unlike Plan-then-Patch agents, which commit to a full DAG before execution and repair it reactively, Evaluate-then-Grow commits one batch at a time, conditioning each expansion on the structured outputs of completed nodes. We next formalize the DAG state, per-node evidence, the Evaluate-then-Grow loop, and dependency-aware context propagation, which together define the execution substrate of DAGent.

\paragraph{DAG State and Node-Level Evidence.}
\label{sec:dag}
A deep research task $\mathcal{T}$ is modeled as a DAG $\mathcal{G}=(V,E)$, where each node $v_i \in V$ is an atomic sub-task and each edge $(v_j, v_i) \in E$ indicates that $v_i$ depends on $v_j$'s output. The graph is initialized empty and incrementally expanded by the Orchestrator. Each node contains a description $\delta_i$ for the sub-task, an execution prompt $p_i$ for the Executor, a dependency set $\operatorname{dep}_i \subset V$ of required upstream nodes, and a status $\sigma_i$ assigned by the Orchestrator. We further define two special node types: a \emph{refine} node, which re-executes a sub-task with an alternative search strategy, and an \emph{answer-type} node, which synthesizes the final response from upstream evidence. After execution, each node outputs a compact \emph{QueryDoc} and preserves a full \emph{InteractionTranscript}.

\paragraph{Per-Node Evidence.} Each node $v_i$ emits a compact structured summary, the QueryDoc ($D_i$):
\begin{equation}
D_i = (\alpha_i,\; \operatorname{expl}_i,\; \operatorname{steps}_i,\; \operatorname{conf}_i,\; \operatorname{unc}_i),
\label{eq:querydoc}
\end{equation}
where $\alpha_i$ is the answer to the current sub-task, $\operatorname{expl}_i$ is a brief explanation of the answer, $\operatorname{steps}_i$ summarizes the key execution steps, $\operatorname{conf}_i \in [0,1]$ denotes the confidence of the Executor, and $\operatorname{unc}_i$ lists uncertainties that may require further verification. The first three fields record the sub-task outcome and rationale, while the last two signal its reliability for Orchestrator evaluation.

In addition, we retain the complete execution trace of each Executor as the InteractionTranscript ($T_i$):
\begin{equation}
T_i = \langle (a_1^{(i)}, o_1^{(i)}),\; \ldots,\; (a_n^{(i)}, o_n^{(i)}) \rangle,
\label{eq:transcript}
\end{equation}
where $a_t^{(i)}$ is an action (e.g., search or page retrieval) at step $t$ and $o_t^{(i)}$ the corresponding observation.

\paragraph{Evaluate-then-Grow DAG Expansion.}
\label{sec:planning}
At each iteration $k$, the Orchestrator $\pi_O$ examines recently completed nodes and produces a planning decision:
\begin{equation}
P_k = \pi_O(\mathcal{T},\; \mathcal{G}_k) = (S_k,\; V_k^{\text{new}}).
\label{eq:planning}
\end{equation}
The state evaluation $S_k$ assigns each node in the latest executed batch a QueryDoc-grounded status: \emph{Success}, indicating confident evidence acquisition; \emph{Uncertain}, indicating evidence that requires further verification or refinement; or \emph{Not Found}, indicating valid execution without relevant evidence. Nodes marked \emph{Uncertain} or \emph{Not Found} spawn refine nodes for further verification or alternative search, with each line capped at three attempts: one initial execution and two refinements. The targeted expansion $V_k^{\text{new}}$ then instantiates new nodes with their descriptions, prompts, and dependency edges; nodes within a batch that share no dependencies execute in parallel. Termination is triggered when the Orchestrator schedules an answer-type node in $V_k^{\text{new}}$; this node depends on the evidence nodes selected as relevant by the Orchestrator and is required to appear alone in its batch, ensuring access to the intended upstream evidence for final synthesis.

\paragraph{Hierarchical Context Propagation.}
\label{sec:context}
Long-horizon deep research can produce InteractionTranscripts of arbitrary length, but downstream nodes typically need only the conclusions of their dependencies. Each sub-task is executed by a ReAct Executor $\pi_R$ equipped with domain-specific tools and \textsc{RecallTool}. Upon completion, $\pi_R$ produces a QueryDoc and preserves its full interaction history as an InteractionTranscript. By default, a node receives only the QueryDocs of its direct dependencies as upstream context. We call this policy \emph{Selective Propagation}, which bounds each node's context by its fan-in rather than the graph depth or total number of nodes. When the compact QueryDoc is insufficient, e.g., when downstream execution needs the exact wording of a passage or a previously rejected candidate, the Executor can retrieve targeted information from the full transcript of a direct dependency by calling the \textsc{RecallTool}:
\begin{equation}
E = \operatorname{Recall}(v_j,\; \textit{goal},\; T_j),
\label{eq:recall}
\end{equation}
where $v_j \in \operatorname{dep}_i$ is a direct dependency, $\textit{goal}$ specifies the information needed by the Executor, and a language model extracts from $T_j$ a focused evidence snippet $E$ conditioned on this goal. \textsc{RecallTool} is a fallback rather than a default: in the Qwen3-32B runs it is invoked in 22.0 / 13.6 / 11.0\% of tasks on BrowseComp-Plus / GAIA / xbench-DeepSearch (Appendix~\ref{appendix:recall_usage}).

\paragraph{\textit{\underline{Remark}}.} Each iteration appends a new batch of nodes with their dependencies, producing a DAG whose growth history is recorded turn by turn; the structural RL signals of \S\ref{sec:dagrpo} are defined on it.

\subsection{DAGRPO: DAG-Conditioned Reinforcement Learning}
\label{sec:dagrpo}
Evaluate-then-Grow produces training trajectories whose DAG topology is recorded turn by turn. This recorded structure exposes a fact that linear-trajectory recipes cannot use: rollouts at different DAG positions contribute unequally to the final answer. Some Executor rollouts lie on the dependency chain feeding the answer-type node, while others explore alternatives the synthesis never consumes.
An outcome-only GRPO recipe assigns the trajectory-level reward $R(\tau) \in \{0, 1\}$ uniformly across all rollouts and ignores this DAG-induced structure. We instantiate the missing structural information as \underline{DAG}-conditioned \underline{G}roup \underline{R}elative \underline{P}olicy \underline{O}ptimization (DAGRPO). DAGRPO adapts GRPO by shaping the per-rollout reward with two structural signals before group normalization: a \emph{topology-conditioned credit} on ReAct Executor rollouts and a \emph{structural compliance regularization} on Orchestrator turns. DAGRPO builds on the standard GRPO objective, injecting DAG-aware structural signals into reward construction without altering the underlying optimization form.

\paragraph{Learning Objective.}
For a task $\mathcal{T}$, we sample $G$ on-policy trajectories from $\pi_{\theta_{\text{old}}}$; each trajectory $\tau$ comprises one Orchestrator sub-rollout and the ReAct Executor sub-rollouts it spawns, jointly indexed by $i$ with response tokens $y_i$, and is assigned a trajectory-level outcome reward $R(\tau) \in \{0, 1\}$ by an LLM judge that verifies final-answer correctness, as detailed in \S\ref{sec:impl}. The DAGRPO objective is
\begin{equation}
\mathcal{J}_{\text{DAGRPO}}(\theta) \;=\; \mathbb{E}\!\left[\frac{1}{|y_\tau|}\sum_{i,\, t}\min\!\Big(\rho_{i,t}(\theta)\, \hat{A}_i^{(u)},\ \operatorname{clip}\!\big(\rho_{i,t}(\theta),\, 1-\epsilon_{\ell},\, 1+\epsilon_h\big)\, \hat{A}_i^{(u)}\Big)\right],
\label{eq:dagrpo_objective}
\end{equation}
where $u \in \{\text{orch}, \text{exec}\}$ denotes the role and $\rho_{i,t}(\theta)$ is the per-token importance ratio between the current and old policy, $|y_\tau| = \sum_i |y_i|$ is the total response-token count in $\tau$, $\epsilon_h > \epsilon_{\ell}$ gives the DAPO-style asymmetric clip \citep{yu2025dapo}, and the role-separated group-relative advantage is
\begin{equation}
\hat{A}_i^{(u)} \;=\; \frac{\tilde{r}_i^{(u)} - \mu_u}{\sigma_u}, \qquad \tilde{r}_i^{(u)} = \begin{cases}
R(\tau) & u = \text{exec},\ v_i \in S(\tau) \\
\alpha\, R(\tau) & u = \text{exec},\ v_i \notin S(\tau) \\
R(\tau) + r_{\text{proc}}(\tau) & u = \text{orch},
\end{cases}
\label{eq:dagrpo_adv}
\end{equation}
where $(\mu_u,\sigma_u)$ are computed within the corresponding role group: the Orchestrator group contains the $G$ sampled trajectories, while the Executor group contains all retained Executor rollouts spawned by them. Here, $\alpha \in [0,1]$ controls the attenuation of off-chain Executor credit, $S(\tau)$ denotes the answer-inclusive closure, and $r_{\text{proc}}(\tau)\leq 0$ is the structural compliance regularization, both detailed below. We also include a small KL-to-reference loss outside Eq.~\eqref{eq:dagrpo_objective} to stabilize long-horizon agent training. Additional details and the reasons behind these choices are given in Appendix~\ref{sec:rationale}.

\paragraph{Topology-Conditioned Credit.} Let $t_{\text{answer}}$ denote the answer-type node scheduled in the terminating iteration. We define the \emph{answer-inclusive closure} $S(\tau)$ as $t_{\text{answer}}$ together with all of its ancestors in $\mathcal{G}$: the set of ReAct Executor rollouts whose outputs directly or transitively fed the final synthesis. ReAct Executors in $S(\tau)$ retain the full $R(\tau)$ while those outside are attenuated by $\alpha$. When $\tau$ contains no answer-type node, we set $\alpha = 1$ as a fallback and recover the outcome-only GRPO behavior. The closure is a structural proxy for contribution rather than a causal attribution: it is read off the recorded graph without re-execution, and Appendix~\ref{sec:rationale} explains why it cannot be exploited by inflating dependency edges during training.

\paragraph{Structural Compliance Regularization.} Evaluate-then-Grow imposes structural constraints on Orchestrator decisions, requiring each plan to be both syntactically valid and consistent with the incremental execution loop. Specifically, the Orchestrator must assign every node in the previous batch an explicit \emph{Success}, \emph{Uncertain}, or \emph{Not Found} status; expand each refinable \emph{Uncertain} or \emph{Not Found} node with a refine node within the three-attempt budget; schedule the answer-type node as a singleton batch so final synthesis can access all upstream evidence; avoid duplicate sub-task prompts; and output parseable JSON with valid dependency references. Let $N(\tau)$ denote the number of Orchestrator turns in $\tau$ that violate these constraints. We define
$r_{\text{proc}}(\tau)=-\lambda_{\text{proc}}\min(N(\tau),K_{\text{proc}})$,
where $\lambda_{\text{proc}}\in(0,1]$ controls the regularization scale and $K_{\text{proc}}$ caps its magnitude. This term is added to the trajectory-level Orchestrator reward to penalize structurally invalid planning decisions.

\paragraph{\textit{\underline{Remark}}.}
DAGent rests on a single structural fact: incremental planning produces a DAG whose growth history is recorded turn by turn, and that recorded structure is what the rest of the method exploits. The hierarchical context layer uses it to bound each node's per-sub-task footprint by graph topology rather than total trajectory length; DAGRPO uses it to define credit-assignment signals that an outcome-only recipe cannot express. Both rest on the incremental loop of \S\ref{sec:dagent}, and removing it weakens both. The complete DAGent workflow is summarized in Algorithm~\ref{alg:dagent} in Appendix~\ref{appendix:algorithm}.

\section{Experiments}
\label{sec:experiments}

We organize the experiments around four research questions \textit{\textbf{(RQs)}}. \textit{\textbf{RQ1}}: How does DAGent compare with deep research baselines on representative benchmarks in both training-free and training-based settings? \textit{\textbf{RQ2}}: Does the lead replicate across backbone scales, vendors, and a frontier closed-source backbone? \textit{\textbf{RQ3}}: Which planning and DAGRPO structural components drive the gains of DAGent? \textit{\textbf{RQ4}}: How efficient is DAGent compared with the Plan-then-Patch alternative?

\subsection{Experimental Setup}
\label{sec:setup}

\paragraph{Benchmarks.}
\label{sec:datasets}
We evaluate on three deep research benchmarks, each with a fixed retrieval backend that every method shares. BrowseComp-Plus \citep{chen2025browsecomp} pairs the original BrowseComp queries \citep{wei2025browsecomp} with a verified corpus; retrieval is local dense retrieval over this corpus with Qwen3-Embedding-8B, and we adopt the 680/150 train/evaluation split of \citet{sun2025foldagent}, balanced across easy, medium, and hard difficulty. GAIA \citep{mialon2024gaia}, a benchmark for evaluating complex task-solving capabilities, is evaluated primarily on its 103-task text-only validation subset with live Google search through Serper and page extraction through Jina; the full 165-task validation set is used only for the GPT-5 327K comparison in Figure~\ref{fig:sota}, for comparability with prior work, with image, document, and audio inspector tools added to handle multi-modal questions as detailed in Appendix~\ref{appendix:cross_backbone}. The 2505 release of xbench-DeepSearch \citep{chen2025xbench, xbench2025deepsearch} contains 100 Chinese-language tasks and uses the same Serper and Jina stack as GAIA. Checkpoints trained on BrowseComp-Plus transfer zero-shot to the other two benchmarks, following the standard practice of agent RL~\citep{sun2025foldagent}.

\paragraph{Baselines.}
\label{sec:baselines}
We compare against five agent families under the same backbone and tool stack: (1) ReAct Agent \citep{yao2023react} at 32K and 109K context with a pre-overflow warning as a termination control; (2) Summary Agent \citep{hu2025hiagent, wu2025resum, kang2025acon, lu2025supo}, which invokes a summary on overflow with up to 10 sessions; (3) Fold Agent \citep{sun2025foldagent, ye2025agentfold, shao2025foldact}, representative of context-folding methods; (4) Flash-Searcher \citep{qin2026flashsearcher}, a Plan-then-Patch DAG agent reproduced from the authors' release with summary interval 4 to fit the 32K-per-sub-task budget; and (5) FlowSearch \citep{hu2025flowsearch}, a Plan-then-Patch DAG agent evaluated with the authors' released implementation with its Coordinator (the execution-conditioned refiner) enabled, at Qwen3-32B only, because its code was released after our initial submission (configuration in Appendix~\ref{appendix:baseline_caps}). At Qwen3-8B we additionally report the three trainable baselines under outcome-only GRPO, and DAGent under both outcome-only GRPO with $\alpha = 1.0$ and no structural regularization (denoted \texttt{GRPO-DAGent}) and DAGRPO (denoted \texttt{DAGRPO-DAGent}); the GRPO variant is the same-budget direct baseline for DAGRPO.

\paragraph{Implementation.}
\label{sec:impl}
Backbones are Qwen3-8B, Qwen3-32B, and Qwen3-235B-A22B \citep{yang2025qwen3} for training-free evaluation and Qwen3-8B for RL training, all with thinking disabled and, per sub-task, a 32K-token response budget on top of an 8K-token prompt (Appendix~\ref{appendix:framework_params}). RL is trained on BrowseComp-Plus with the DAPO asymmetric clip \citep{yu2025dapo}, off-chain credit $\alpha=0.5$, and the structural compliance regularization. Inference uses greedy decoding; Pass@1 is scored by a human-calibrated LLM judge, with details in Appendix~\ref{appendix:impl_details}.

\subsection{Main Results (RQ1, RQ2)}
\label{sec:main_results}

\begin{table}[t]
\caption{Main results on BrowseComp-Plus, GAIA, and xbench-DeepSearch: Pass@1 (\%), best result per backbone in \textbf{bold}. Training-free rows use a single greedy inference pass; training-based rows report mean$\pm$std (sample standard deviation) over three training seeds on the benchmark averages. FlowSearch is evaluated with the authors' released implementation at Qwen3-32B only (\S\ref{sec:baselines}).}
\label{tab:main}
\centering
\scriptsize
\setlength{\tabcolsep}{3pt}
\resizebox{\linewidth}{!}{%
\begin{tabular}{llllcccccccc}
\toprule
\multirow{2}{*}{\textbf{Backbone}} & \multirow{2}{*}{\textbf{Max \#Token}} & \multirow{2}{*}{\textbf{Agent Paradigm}} & \multicolumn{4}{c}{\textbf{BrowseComp-Plus}} & \multicolumn{4}{c}{\textbf{GAIA}} & \textbf{xbench-DS} \\
\cmidrule(lr){4-7} \cmidrule(lr){8-11}
 & & & Easy & Med. & Hard & Avg. & L1 & L2 & L3 & Avg. & Avg. \\
\midrule
\rowcolor{blue!8}
\multicolumn{12}{c}{\textbf{Training-free}} \\
\midrule
\multirow{6}{*}{Qwen3-8B}
 & 32K          & ReAct Agent                       & 46.0 & 14.0 & 0.0 & 20.0 & 38.5 & 23.1 & 25.0 & 29.1 & 39.0 \\
 & 109K         & ReAct Agent                       & 64.0 & 18.0 & 2.0 & 28.0 & 46.2 & 26.9 & 25.0 & 34.0 & 44.0 \\
 & 32K$\times$N & Summary Agent                     & 76.0 & 20.0 & 2.0 & 32.7 & 51.3 & 26.9 & 16.7 & 35.0 & 54.0 \\
 & 32K$\times$N & Fold Agent                        & 78.0 & 24.0 & 2.0 & 34.7 & 48.7 & 28.8 & 16.7 & 35.0 & 52.0 \\
 & 32K$\times$N & Flash-Searcher                    & 78.0 & 24.0 & 2.0 & 34.7 & 51.3 & 32.7 & 25.0 & 38.8 & 55.0 \\
 \rowcolor{yellow!10}
 & 32K$\times$N & \textbf{DAGent (Ours)}            & \textbf{84.0} & \textbf{30.0} & \textbf{6.0} & \textbf{40.0} & \textbf{59.0} & \textbf{40.4} & \textbf{33.3} & \textbf{46.6} & \textbf{60.0} \\
\midrule
\multirow{7}{*}{Qwen3-32B}
 & 32K          & ReAct Agent                       & 50.0 & 16.0 & 2.0 & 22.7 & 46.2 & 28.8 & 8.3 & 33.0 & 54.0 \\
 & 109K         & ReAct Agent                       & 66.0 & 24.0 & 4.0 & 31.3 & 51.3 & 36.5 & 16.7 & 39.8 & 58.0 \\
 & 32K$\times$N & Summary Agent                     & 80.0 & 36.0 & 0.0 & 38.7 & 51.3 & 28.8 & 25.0 & 36.9 & 55.0 \\
 & 32K$\times$N & Fold Agent                        & 82.0 & 40.0 & 2.0 & 41.3 & 56.4 & 38.5 & 16.7 & 42.7 & 55.0 \\
 & 32K$\times$N & Flash-Searcher                    & 80.0 & 34.0 & 2.0 & 38.7 & 56.4 & 40.4 & 25.0 & 44.7 & 58.0 \\
 & 32K$\times$N & FlowSearch                        & 84.0 & 42.0 & 4.0 & 43.3 & 66.7 & 46.2 & 25.0 & 51.5 & 63.0 \\
 \rowcolor{yellow!10}
 & 32K$\times$N & \textbf{DAGent (Ours)}            & \textbf{88.0} & \textbf{46.0} & \textbf{8.0} & \textbf{47.3} & \textbf{69.2} & \textbf{50.0} & \textbf{33.3} & \textbf{55.3} & \textbf{65.0} \\
\midrule
\multirow{6}{*}{Qwen3-235B}
 & 32K          & ReAct Agent                       & 84.0 & 32.0 & 10.0 & 42.0 & 56.4 & 55.8 & 33.3 & 53.4 & 60.0 \\
 & 109K         & ReAct Agent                       & 90.0 & 50.0 & 16.0 & 52.0 & 61.5 & 57.7 & \textbf{41.7} & 57.3 & 64.0 \\
 & 32K$\times$N & Summary Agent                     & 88.0 & 60.0 & 20.0 & 56.0 & 59.0 & 53.8 & 25.0 & 52.4 & 60.0 \\
 & 32K$\times$N & Fold Agent                        & \textbf{94.0} & 56.0 & 14.0 & 54.7 & 61.5 & 53.8 & 25.0 & 53.4 & 62.0 \\
 & 32K$\times$N & Flash-Searcher                    & 90.0 & 54.0 & 14.0 & 52.7 & 64.1 & 55.8 & 25.0 & 55.3 & 70.0 \\
 \rowcolor{yellow!10}
 & 32K$\times$N & \textbf{DAGent (Ours)}            & \textbf{94.0} & \textbf{68.0} & \textbf{22.0} & \textbf{61.3} & \textbf{74.4} & \textbf{59.6} & \textbf{41.7} & \textbf{63.1} & \textbf{72.0} \\
\midrule
\rowcolor{blue!8}
\multicolumn{12}{c}{\textbf{Training-based}} \\
\midrule
\multirow{4}{*}{Qwen3-8B}
 & 32K          & GRPO-ReAct Agent          & 72.0 & 16.0 & 4.0 & $30.7 \pm 1.8$ & 41.0 & 26.9 & 25.0 & $32.0 \pm 2.9$ & $43.0 \pm 2.6$ \\
 & 109K         & GRPO-ReAct Agent          & 76.0 & 24.0 & 4.0 & $34.7 \pm 2.4$ & 48.7 & 30.8 & 25.0 & $36.9 \pm 1.9$ & $48.0 \pm 2.0$ \\
 & 32K$\times$N & GRPO-Summary Agent        & 80.0 & 30.0 & 4.0 & $38.0 \pm 2.0$ & 53.8 & 30.8 & 25.0 & $38.8 \pm 1.7$ & $57.0 \pm 1.7$ \\
 & 32K$\times$N & GRPO-Fold Agent           & 84.0 & 34.0 & 6.0 & $41.3 \pm 1.3$ & 51.3 & 32.7 & 25.0 & $38.8 \pm 2.6$ & $56.0 \pm 3.0$ \\
 \midrule
 \rowcolor{yellow!5}
 Qwen3-8B
 & 32K$\times$N & \textbf{GRPO-DAGent}      & 88.0 & 40.0 & 10.0 & $46.0 \pm 1.2$ & 61.5 & 46.8 & \textbf{33.3} & $50.8 \pm 1.1$ & $63.0 \pm 1.0$ \\
 \rowcolor{yellow!10}
 Qwen3-8B
 & 32K$\times$N & \textbf{DAGRPO-DAGent}    & \textbf{90.0} & \textbf{48.0} & \textbf{10.7} & $\mathbf{49.6 \pm 1.0}$ & \textbf{64.1} & \textbf{50.0} & \textbf{33.3} & $\mathbf{53.4 \pm 1.0}$ & $\mathbf{65.7 \pm 1.5}$ \\
\bottomrule
\end{tabular}%
}
\end{table}

Table~\ref{tab:main} reports Pass@1 across three benchmarks for DAGent and the baselines of \S\ref{sec:baselines} at three Qwen3 backbone scales, in training-free and training-based settings.

\begin{wrapfigure}[14]{r}{0.5\linewidth}
  \centering
  \includegraphics[width=\linewidth]{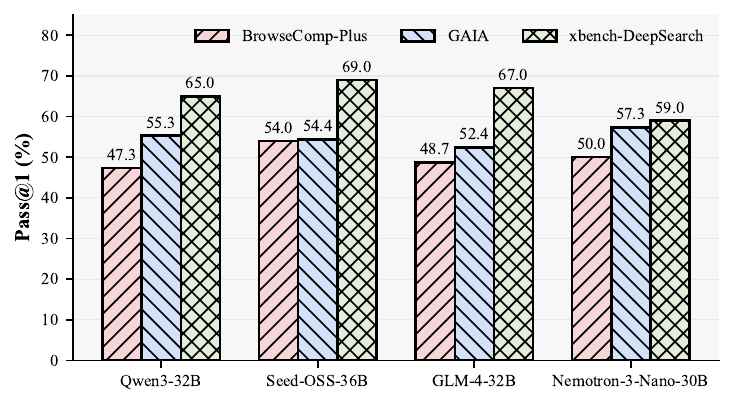}
  \caption{Cross-backbone DAGent Pass@1 (\%); per-baseline breakdown in Appendix~\ref{appendix:cross_backbone}.}
  \label{fig:cross_backbone}
\end{wrapfigure}

\obs{1}{DAGent leads the strongest non-DAGent training-free baseline at every Qwen3 backbone scale across the three benchmarks.}
At Qwen3-32B, DAGent reaches 47.3 / 55.3 / 65.0 on BrowseComp-Plus / GAIA / xbench-DeepSearch, 4.0 / 3.8 / 2.0 points above FlowSearch, the strongest baseline at this scale, and 6.0 / 10.6 / 7.0 points above the strongest of the remaining baselines. The lead persists at 8B (+5.3 / +7.8 / +5.0) and 235B-A22B (+5.3 / +5.8 / +2.0), and on every difficulty split DAGent matches or exceeds every baseline, with two ties at 235B-A22B; relative gains grow with difficulty at 8B and 32B.

\obs{2}{DAGRPO further improves Qwen3-8B DAGent over a same-budget outcome-only GRPO baseline by 3.0 average points, isolating the gain to its structural signals.}
DAGRPO improves the BrowseComp-Plus / GAIA / xbench-DeepSearch Pass@1 of DAGent from 40.0 / 46.6 / 60.0 in training-free mode to 49.6 / 53.4 / 65.7 over three seeds, exceeding the GRPO baseline by 3.6 / 2.6 / 2.7 points; per-seed values appear in Appendix~\ref{appendix:per_seed}. The same outcome-only GRPO recipe improves each trainable baseline by 2.9 to 10.7 points across the three benchmarks, so the further DAGRPO gain on DAGent localizes to the structural signals rather than the training budget.

\begin{figure}[!ht]
\centering
\includegraphics[width=\linewidth]{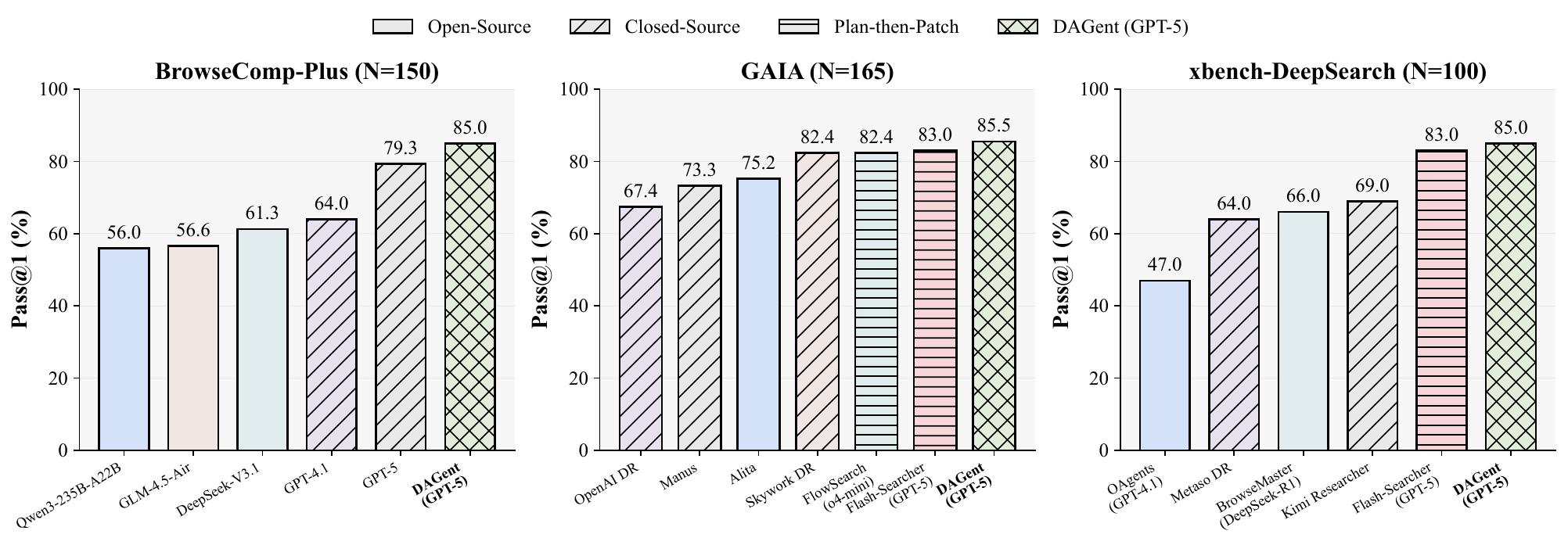}
\caption{Performance comparison of DAGent and state-of-the-art agent systems on BrowseComp-Plus, GAIA, and xbench-DeepSearch. Pass@1 (\%) is reported.}
\label{fig:sota}
\end{figure}

\obs{3}{The lead replicates across four mid-size open-source backbones from four vendors and persists at frontier scale with GPT-5 at 327K context.}
Figure~\ref{fig:cross_backbone} shows that DAGent maintains the per-backbone top position across Qwen3-32B, Seed-OSS-36B \citep{seed2025seedoss}, GLM-4-32B \citep{glm2025glm432b}, and Nemotron-3-Nano-30B \citep{nvidia2025nemotron3}, with the per-baseline breakdown in Appendix~\ref{appendix:cross_backbone}. DAGent at GPT-5 327K leads the strongest reported baseline by 5.7 / 2.5 / 2.0 points on BrowseComp-Plus / GAIA / xbench-DeepSearch (Figure~\ref{fig:sota}), with GAIA evaluated on its full 165-task validation set with multi-modal tools; the baseline list and number provenance appear in Appendix~\ref{appendix:cross_backbone}.

\begin{table}[!ht]
\caption{Component ablation with Qwen3-32B in training-free mode: Pass@1 (\%) per benchmark, Overall is the unweighted mean, and $\Delta$ is the change from DAGent (Full). All rows use a single greedy inference pass, as in the training-free rows of Table~\ref{tab:main}.}
\label{tab:ablation}
\centering
\small
\setlength{\tabcolsep}{4pt}
\resizebox{\linewidth}{!}{%
\begin{tabular}{@{}l*{8}{c}@{}}
\toprule
\multirow{2}{*}{\textbf{Method}} & \multicolumn{2}{c}{\textbf{BrowseComp-Plus}} & \multicolumn{2}{c}{\textbf{GAIA}} & \multicolumn{2}{c}{\textbf{xbench-DS}} & \multicolumn{2}{c}{\textbf{Overall}} \\
\cmidrule(lr){2-3}\cmidrule(lr){4-5}\cmidrule(lr){6-7}\cmidrule(lr){8-9}
& Avg. & $\Delta$ & Avg. & $\Delta$ & Avg. & $\Delta$ & Avg. & $\Delta$ \\
\midrule
 \rowcolor{yellow!10}
\textbf{DAGent (Full)}                & \textbf{47.3} & --- & \textbf{55.3} & --- & \textbf{65.0} & --- & \textbf{55.9} & --- \\
\quad Plan-then-Patch Variant         & 42.0 & \textbf{\textcolor[RGB]{0,128,128}{-5.3}}  & 50.5 & \textbf{\textcolor[RGB]{0,128,128}{-4.8}}  & 61.0 & \textbf{\textcolor[RGB]{0,128,128}{-4.0}}  & 51.2 & \textbf{\textcolor[RGB]{0,128,128}{-4.7}} \\
\quad w/o Evaluate-then-Grow Planning & 33.3 & \textbf{\textcolor[RGB]{0,128,128}{-14.0}} & 42.7 & \textbf{\textcolor[RGB]{0,128,128}{-12.6}} & 55.0 & \textbf{\textcolor[RGB]{0,128,128}{-10.0}} & 43.7 & \textbf{\textcolor[RGB]{0,128,128}{-12.2}} \\
\quad w/o Selective Propagation       & 34.7 & \textbf{\textcolor[RGB]{0,128,128}{-12.6}} & 44.7 & \textbf{\textcolor[RGB]{0,128,128}{-10.6}} & 57.0 & \textbf{\textcolor[RGB]{0,128,128}{-8.0}}  & 45.5 & \textbf{\textcolor[RGB]{0,128,128}{-10.4}} \\
\quad w/o QueryDoc                    & 39.3 & \textbf{\textcolor[RGB]{0,128,128}{-8.0}}  & 47.6 & \textbf{\textcolor[RGB]{0,128,128}{-7.7}}  & 60.0 & \textbf{\textcolor[RGB]{0,128,128}{-5.0}}  & 49.0 & \textbf{\textcolor[RGB]{0,128,128}{-6.9}} \\
\quad w/o InteractionTranscript       & 43.3 & \textbf{\textcolor[RGB]{0,128,128}{-4.0}}  & 52.4 & \textbf{\textcolor[RGB]{0,128,128}{-2.9}}  & 63.0 & \textbf{\textcolor[RGB]{0,128,128}{-2.0}}  & 52.9 & \textbf{\textcolor[RGB]{0,128,128}{-3.0}} \\
\bottomrule
\end{tabular}%
}
\end{table}

\subsection{Ablation Study (RQ3)}
\label{sec:ablation}

This subsection ablates DAGent at two layers: Table~\ref{tab:ablation} dissects the prompt layer on Qwen3-32B, and Table~\ref{tab:dagrpo_analysis} dissects the RL layer on Qwen3-8B, where the DAGRPO and GRPO-baseline rows reproduce the three-seed means of Table~\ref{tab:main} and the remaining rows use a single training seed.

\obs{4}{Within the same DAGent architecture, evidence-conditioned incremental graph growth outperforms pre-execution commitment with reactive patches, which in turn outperforms a single upfront plan; the three context components contribute non-trivially on top.}
Replacing Evaluate-then-Grow with Plan-then-Patch costs 5.3 / 4.8 / 4.0 points on BrowseComp-Plus / GAIA / xbench-DeepSearch, isolating evidence-conditioned planning from the generic patch capability; collapsing further to a single upfront plan with no replanning costs another 8.7 / 7.8 / 6.0 points, totaling 14.0 / 12.6 / 10.0. Among context-management components the drop ordering is consistent across the three benchmarks: Selective Propagation (12.6 / 10.6 / 8.0) outranks QueryDoc (8.0 / 7.7 / 5.0), which outranks InteractionTranscript (4.0 / 2.9 / 2.0); the last row is also the \textsc{RecallTool} ablation, since removing the transcript disables recall (Appendix~\ref{appendix:recall_usage}).

\obs{5}{Both DAGRPO structural signals contribute positive Pass@1 gains, with the off-chain credit coefficient $\alpha$ peaking at 0.5.}
Removing topology-conditioned credit from full DAGRPO costs 2.3 / 1.9 / 1.7 points on the three benchmarks, and removing the structural compliance regularization costs 1.6 / 1.0 / 0.7 points (Table~\ref{tab:dagrpo_analysis}). Because the ablation rows use a single seed while the DAGRPO row is a three-seed mean with std 1.0 / 1.0 / 1.5, the smaller per-benchmark differences are within seed noise; the support for both signals is the consistent sign of every per-benchmark difference and the Overall differences of $-2.0$ and $-1.1$. The $\alpha$ sweep peaks at 0.5 on every benchmark; $\alpha = 0$ pushes performance below the GRPO baseline as off-chain learning collapses, while $\alpha = 0.25$ retains 75 / 62 / 37\% of the gain over the GRPO baseline because partial attenuation already polarizes the credit signal. Training-time mechanism traces of both signals appear in Appendix~\ref{appendix:dagrpo_dynamics}.

\begin{table}[!ht]
\caption{DAGRPO component ablation and $\alpha$ sensitivity on Qwen3-8B: Pass@1 (\%) per benchmark, Overall is the unweighted mean, and $\Delta$ is the change from DAGRPO; the Overall $\Delta$ is the mean of the three per-benchmark $\Delta$ values and can differ by 0.1 from the difference of the rounded Overall values. The DAGRPO and GRPO-baseline rows are the three-seed means of Table~\ref{tab:main}; all other rows are single-seed runs with the same training budget.}
\label{tab:dagrpo_analysis}
\centering
\small
\setlength{\tabcolsep}{4pt}
\resizebox{\linewidth}{!}{%
\begin{tabular}{@{}ccl*{8}{c}@{}}
\toprule
\multirow{2}{*}{$\alpha$} & \multirow{2}{*}{\textbf{Regularization}} & \multirow{2}{*}{\textbf{Configuration}} & \multicolumn{2}{c}{\textbf{BrowseComp-Plus}} & \multicolumn{2}{c}{\textbf{GAIA}} & \multicolumn{2}{c}{\textbf{xbench-DS}} & \multicolumn{2}{c}{\textbf{Overall}} \\
\cmidrule(lr){4-5}\cmidrule(lr){6-7}\cmidrule(lr){8-9}\cmidrule(lr){10-11}
& & & Avg. & $\Delta$ & Avg. & $\Delta$ & Avg. & $\Delta$ & Avg. & $\Delta$ \\
\midrule
0.00 & on  & ---                              & 44.0 & \textbf{\textcolor[RGB]{0,128,128}{-5.6}} & 49.5 & \textbf{\textcolor[RGB]{0,128,128}{-3.9}} & 61.0 & \textbf{\textcolor[RGB]{0,128,128}{-4.7}} & 51.5 & \textbf{\textcolor[RGB]{0,128,128}{-4.7}} \\
0.25 & on  & ---                              & 48.7 & \textbf{\textcolor[RGB]{0,128,128}{-0.9}} & 52.4 & \textbf{\textcolor[RGB]{0,128,128}{-1.0}} & 64.0 & \textbf{\textcolor[RGB]{0,128,128}{-1.7}} & 55.0 & \textbf{\textcolor[RGB]{0,128,128}{-1.2}} \\
\rowcolor{yellow!10}
0.50 & on  & \textbf{DAGRPO}                  & \textbf{49.6} & --- & \textbf{53.4} & --- & \textbf{65.7} & --- & \textbf{56.2} & --- \\
0.75 & on  & ---                              & 48.0 & \textbf{\textcolor[RGB]{0,128,128}{-1.6}} & 52.4 & \textbf{\textcolor[RGB]{0,128,128}{-1.0}} & 64.0 & \textbf{\textcolor[RGB]{0,128,128}{-1.7}} & 54.8 & \textbf{\textcolor[RGB]{0,128,128}{-1.4}} \\
1.00 & on  & w/o topology credit              & 47.3 & \textbf{\textcolor[RGB]{0,128,128}{-2.3}} & 51.5 & \textbf{\textcolor[RGB]{0,128,128}{-1.9}} & 64.0 & \textbf{\textcolor[RGB]{0,128,128}{-1.7}} & 54.3 & \textbf{\textcolor[RGB]{0,128,128}{-2.0}} \\
\midrule
0.50 & off & w/o compliance regularization    & 48.0 & \textbf{\textcolor[RGB]{0,128,128}{-1.6}} & 52.4 & \textbf{\textcolor[RGB]{0,128,128}{-1.0}} & 65.0 & \textbf{\textcolor[RGB]{0,128,128}{-0.7}} & 55.1 & \textbf{\textcolor[RGB]{0,128,128}{-1.1}} \\
1.00 & off & GRPO baseline                    & 46.0 & \textbf{\textcolor[RGB]{0,128,128}{-3.6}} & 50.8 & \textbf{\textcolor[RGB]{0,128,128}{-2.6}} & 63.0 & \textbf{\textcolor[RGB]{0,128,128}{-2.7}} & 53.3 & \textbf{\textcolor[RGB]{0,128,128}{-3.0}} \\
\bottomrule
\end{tabular}%
}
\end{table}

\subsection{Efficiency Analysis (RQ4)}
\label{sec:efficiency}

We measure the per-task footprint of DAGent with Qwen3-32B in training-free mode. The 32K$\times$N of Table~\ref{tab:main} is a per-sub-task cap shared by all multi-context methods, not a per-task budget; the totals below are what each workflow spends under it. Execution steps count Orchestrator rounds plus the longest Executor trajectory per parallel batch; tool calls count all search, open\_page, recall, and plan invocations. Appendix~\ref{appendix:efficiency_extended} extends the comparison to Flash-Searcher and FlowSearch and adds external tool calls (search and open\_page only) and wall-clock time.

\begin{figure}[!ht]
\centering
\includegraphics[width=\linewidth]{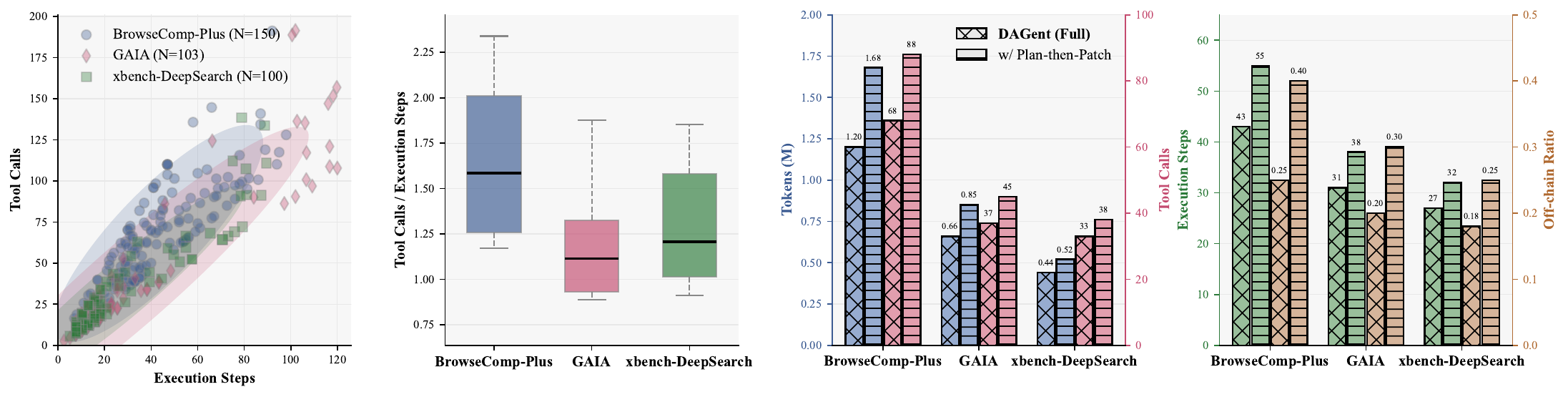}\\[-2pt]
\makebox[0.489\linewidth][c]{\small(a)}\makebox[0.511\linewidth][c]{\small(b)}
\caption{DAGent per-task efficiency on Qwen3-32B training-free. (a) Tool calls vs. execution steps with 2.2$\sigma$ confidence ellipses, and per-task tool calls per step. (b) DAGent (Full) vs. the Plan-then-Patch variant on per-task footprint metrics.}
\label{fig:efficiency}
\end{figure}

\obs{6}{The per-task tool-call-to-step ratio of DAGent is tightly distributed and scales with task complexity, so the DAG-based architecture adds tool calls in proportion to task difficulty rather than inflating redundant steps.}
Median tool-call-to-step ratios are 1.6, 1.1, and 1.2 on BrowseComp-Plus, GAIA, and xbench-DeepSearch in Figure~\ref{fig:efficiency}(a), tracking task complexity, with a tight interquartile range on GAIA. Mean per-task footprints are 42.7 / 31.2 / 26.8 steps, 67.6 / 36.8 / 32.9 tool calls, and 1.20M / 0.66M / 0.44M total input plus output tokens.

\obs{7}{Within the same DAGent architecture, Plan-then-Patch inflates per-task token, tool-call, and node footprints while accuracy lags Evaluate-then-Grow on every benchmark.}
Plan-then-Patch commits a broader DAG before observing node-level evidence, increasing the off-chain Executor ratio from 0.25 / 0.20 / 0.18 to 0.40 / 0.30 / 0.25 and adding 40 / 29 / 18\% more tokens plus 28 / 21 / 21\% more execution steps (Figure~\ref{fig:efficiency}(b); per-task means in Table~\ref{tab:efficiency_extended} of Appendix~\ref{appendix:efficiency_extended}). It also makes 35 / 27 / 21\% more external tool calls and needs 35 / 27 / 23\% more wall-clock time per task, yet still trails DAGent (Full) by 5.3 / 4.8 / 4.0 Pass@1 points (Table~\ref{tab:ablation}): incremental, evidence-conditioned planning avoids redundant branches while improving accuracy.

\section{Conclusion}
\label{sec:conclusion}
We presented DAGent, a DAG-based multi-agent framework for deep research that replaces Plan-then-Patch commitment with Evaluate-then-Grow incremental planning, and DAGRPO, a DAG-conditioned reinforcement learning adaptation that exploits the recorded topology Evaluate-then-Grow produces. Across BrowseComp-Plus, GAIA, and xbench-DeepSearch, DAGent consistently outperforms strong open-source baselines across backbone scales, vendors, and frontier-scale settings, leads the authors' released FlowSearch implementation in a controlled Qwen3-32B comparison, and does so at a lower per-task cost than its Plan-then-Patch counterparts. DAGRPO adds further gains over a same-budget outcome-only GRPO baseline.

\bibliographystyle{plainnat}
\bibliography{references}

\clearpage
\appendix
\part{Appendix}
\parttoc

\section{DAGent Algorithm}
\label{appendix:algorithm}

Algorithm~\ref{alg:dagent} summarizes the complete DAGent workflow described in \S\ref{sec:dagent}.

\begin{algorithm}[!ht]
\caption{DAGent: Evaluate-then-Grow Workflow}
\label{alg:dagent}
\begin{algorithmic}[1]
\REQUIRE Task $\mathcal{T}$, Orchestrator policy $\pi_O$, ReAct Executor policy $\pi_R$, maximum iterations $K$
\ENSURE Final answer $\hat{a}$
\STATE $\mathcal{G} \leftarrow (\emptyset, \emptyset)$ \hfill $\triangleright$ Initialize empty DAG
\FOR{$k = 1, 2, \ldots, K$}
    \STATE $(S_k, V_k^{\text{new}}) \leftarrow \pi_O(\mathcal{T}, \mathcal{G})$ \hfill $\triangleright$ Evaluate \& plan, Eq.~\eqref{eq:planning}
    \STATE Update node statuses in $\mathcal{G}$ according to $S_k$
    \STATE Add nodes $V_k^{\text{new}}$ and dependency edges to $\mathcal{G}$
    \FOR{each $v_i \in V_k^{\text{new}}$ \textbf{in parallel}}
        \STATE $C_i \leftarrow \bigoplus_{j \in \operatorname{dep}_i} D_j$ \hfill $\triangleright$ Selective Propagation
        \STATE $(D_i, T_i) \leftarrow \pi_R(p_i, C_i)$ \hfill $\triangleright$ Sub-task execution
        \STATE Store $D_i, T_i$ in node $v_i$
    \ENDFOR
    \IF{$V_k^{\text{new}}$ contains an answer-type node}
        \STATE \textbf{break}
    \ENDIF
\ENDFOR
\RETURN answer field from $D_{t_{\text{answer}}}$
\end{algorithmic}
\end{algorithm}

\section{Design Decisions}
\label{sec:rationale}

This appendix explains the main design decisions of DAGent and DAGRPO and relates DAGRPO to concurrent credit-assignment methods.

\subsection{Evaluate-then-Grow versus Plan-then-Patch}
\label{appendix:rationale_planning}

The Plan-then-Patch label refers to when a planning decision is made and how much the system knows at that moment, not to how many nodes one call produces. Flash-Searcher \citep{qin2026flashsearcher} produces its entire plan in one decomposition call before execution and then periodically updates it, removing resolved nodes and inserting new ones. FlowSearch \citep{hu2025flowsearch} builds its knowledge flow over several planner iterations that all complete before any node runs, and then adjusts the flow with graph transformation operations based on intermediate outcomes. In both systems a plan that covers the whole task exists before execution, and execution-time edits patch that standing plan. In DAGent every node, including the answer node, is created only after the nodes it depends on have executed and been evaluated, and no node is ever deleted or rewired, so the recorded graph is exactly the graph that ran. The first batch is the one decision made without completed nodes: it is planned from the task description alone and kept small, and Appendix~\ref{appendix:first_batch} reports how often a first-batch node succeeds and how often a weak one is repaired by a refine node, where a node counts as weak when it has been evaluated as \emph{Uncertain} or \emph{Not Found}.

\subsection{DAGRPO}
\label{appendix:rationale_dagrpo}

\paragraph{Role-separated normalization.}
The Orchestrator and the Executors produce different numbers of rollouts per task and operate on different reward distributions. A single shared $(\mu, \sigma)$ across both roles would let the more numerous Executor rewards dominate the Orchestrator gradient signal, so the group statistics in Eq.~\eqref{eq:dagrpo_adv} are computed within each role.

\paragraph{Multiplicative off-chain attenuation.}
The form $\tilde{r}_i^{(\text{exec, off-chain})} = \alpha R(\tau)$ is chosen because it gives no topology signal on failed trajectories with $R(\tau) = 0$. Off-chain rollouts in a failed trajectory may carry useful evidence that the synthesis never integrated, and a constant subtractive penalty would push the policy away from such queries. The $\alpha$ sweep in Table~\ref{tab:dagrpo_analysis} supports this reading: removing all off-chain credit ($\alpha = 0$) falls below the outcome-only GRPO baseline, so off-chain rollouts carry learning value, and $\alpha = 0.5$ is the best value on every benchmark.

\paragraph{The answer-inclusive closure as a structural proxy.}
$S(\tau)$ is a reachability signal read off the recorded graph without re-execution. It is a proxy for contribution rather than a causal attribution, and three properties prevent it from being exploited by inflating dependencies during training. First, nothing in the objective grows with the number of ancestors: the Orchestrator reward $R(\tau) + r_{\text{proc}}(\tau)$ does not count them, and Executor advantages are centered on the group mean, so moving more rollouts on-chain only redistributes credit. If every rollout were on-chain, the objective would reduce to the $\alpha = 1.0$ row with regularization on in Table~\ref{tab:dagrpo_analysis}, which DAGRPO beats by 2.0 points overall. Second, every extra edge costs accuracy: the DAG is append-only, so the closure can only be enlarged by adding a dependency edge, and under Selective Propagation each such edge places an irrelevant QueryDoc into the context of the node that consumes it. Third, the training runs show the opposite trend: dependency inflation would enlarge the graph and drive the off-chain ratio toward zero, whereas Appendix~\ref{appendix:rl_efficiency} shows the node count per task falling, the off-chain ratio falling but staying well above zero, and Pass@1 rising under both GRPO and DAGRPO.

\paragraph{Trajectory-level structural penalty.}
The scale $\lambda_{\text{proc}} = 0.3$ keeps the penalty of a correct plan with up to three violating turns below the outcome reward, so such a plan is not ranked below a clean failure, and the cap $K_{\text{proc}} = 5$ bounds the penalty of a heavily violating plan at $1.5$. Aggregating $r_{\text{proc}}$ at the trajectory level rather than at the token level keeps the Orchestrator gradient tied to $R(\tau)$ and lets the group baseline absorb the across-trajectory mean of frequent violations. Apart from the requirement of parseable JSON without duplicate sub-task prompts, the constraints that the penalty enforces (\S\ref{sec:dagrpo}) control how evidence flows rather than how a plan looks: every executed node must receive an explicit status, every weak node with attempts left must be re-examined, the final synthesis must see all evidence assigned to it, and dependencies may point only to nodes that have already run and been evaluated. Appendix~\ref{appendix:violations} reports the association between structural violations and task accuracy observed in the evaluation logs.

\subsection{Relation to Concurrent Credit-Assignment Methods}
\label{appendix:concurrent_rl}

Three concurrent methods are closest to DAGRPO. Graph-GRPO \citep{cang2026graphgrpo} optimizes the communication topology over a fixed set of agents by scoring each edge across a sampled group of topologies; in DAGent the graph is the task decomposition itself, and its nodes do not exist until the run creates them. M-GRPO \citep{hong2025mgrpo} computes group-relative advantages separately for a main agent and its sub-agents, which is the closest analogue of our role-separated normalization, but its credit depends only on the role of a rollout, whereas ours also depends on whether the rollout feeds the answer node. CCPO \citep{li2026ccpo} estimates the marginal contribution of an agent by removing it counterfactually; in a research DAG the nodes below a removed node have already consumed its QueryDoc, so the counterfactual would require re-executing that subgraph, whereas $S(\tau)$ is read off the recorded graph without re-execution. Append-only growth is what makes this signal well defined: the ancestry of a node in the final graph is the context it saw when it ran, which no longer holds once a refiner can delete or rewire nodes after they have executed, as in FlowSearch and Flash-Searcher.

\section{Implementation Details}
\label{appendix:impl_details}

\subsection{Benchmarks, Tool Stack, and Judge}
\label{appendix:tool_stack}

Table~\ref{tab:impl_benchmarks} lists the evaluation set, retrieval backend, and judge prompt of each benchmark. Within a benchmark every method uses the same backend, so differences between methods come from the workflow and not from the retrieval API. Two of the three benchmarks use live Google search; BrowseComp-Plus uses local dense retrieval with Qwen3-Embedding-8B over its fixed corpus because its protocol requires it. Pass@1 is scored by GPT-4o-mini with GPT-4.1 as a tie-breaker on borderline cases (Appendix~\ref{appendix:human_agreement}); the judge prompts are listed in Appendix~\ref{appendix:prompts}.

\begin{table}[!ht]
\caption{Benchmarks and tool stack; all methods on a benchmark share its backend and judge.}
\label{tab:impl_benchmarks}
\centering
\small
\setlength{\tabcolsep}{5pt}
\begin{tabular}{@{}lccc@{}}
\toprule
 & \textbf{BrowseComp-Plus} & \textbf{GAIA} & \textbf{xbench-DeepSearch} \\
\midrule
Tasks            & 150                           & 103 (text-only validation) & 100 (2505 release) \\
Difficulty splits & 50 easy / 50 medium / 50 hard & 39 L1 / 52 L2 / 12 L3     & none \\
Language         & English                       & English                    & Chinese \\
Search           & Local dense retrieval         & Google search via Serper   & Google search via Serper \\
Page access      & Verified corpus               & Jina extraction            & Jina extraction \\
Judge prompt     & Benchmark rubric              & Equivalence prompt         & Benchmark rubric (Chinese) \\
\bottomrule
\end{tabular}
\end{table}

\subsection{DAGent Framework Parameters}
\label{appendix:framework_params}

Table~\ref{tab:impl_framework} lists the parameters of the DAGent framework. The context budget is enforced per agent context: each Orchestrator turn and each Executor sub-task runs in its own context whose prompt is capped at 8,192 tokens and whose response, including tool outputs, is capped at 32,768 tokens. This is the 32K$\times$N budget of Table~\ref{tab:main}, which every multi-context baseline shares. The Executor timeout applies to a single sub-task; the whole task is additionally bounded by a session timeout of 5,400\,s, a bound that no evaluated task reaches.

\begin{table}[!ht]
\caption{DAGent framework parameters, grouped by scope. ``Per context'' denotes one Orchestrator turn or one Executor sub-task; ``per task'' denotes one benchmark query.}
\label{tab:impl_framework}
\centering
\small
\begin{tabular}{@{}lll@{}}
\toprule
\textbf{Parameter} & \textbf{Scope} & \textbf{Value} \\
\midrule
Prompt length                                     & per context      & $8{,}192$ tokens \\
Response length, including tool outputs           & per context      & $32{,}768$ tokens \\
\addlinespace
Orchestrator iterations $K$                       & per task         & $30$ \\
Session timeout                                   & per task         & $5{,}400$\,s \\
\addlinespace
Executor turns                                    & per sub-task     & $200$ \\
Executor timeout, evaluation / training           & per sub-task     & $600$\,s / $1{,}800$\,s \\
Attempts, one initial execution plus refines      & per sub-task     & $3$ \\
\addlinespace
Search results top-$k$                            & per search call  & $10$ \\
Decoding                                          & --               & greedy, thinking disabled \\
\bottomrule
\end{tabular}
\end{table}

\subsection{Baseline Configurations and Caps}
\label{appendix:baseline_caps}

Table~\ref{tab:impl_caps} lists the budget-related settings of the baseline workflows; the DAGent settings are those of Table~\ref{tab:impl_framework}, and the Plan-then-Patch variant shares them and adds a patch budget. Flash-Searcher keeps the 40-step cap of the authors' release; only its summary interval is changed from 8 to 4 so that each summary fits within the 32K per-sub-task budget shared by all multi-context methods. FlowSearch runs the official question-answering preset of the released implementation with three documented changes: the model of every role is set to the evaluation backbone, the Coordinator (the execution-conditioned refiner described by \citet{hu2025flowsearch}) is enabled because the preset ships with it disabled, and the tool set is restricted to the shared \texttt{search} and \texttt{open\_page} backends; all official budgets are kept. The effective configuration is written into every result record.

\begin{table}[!ht]
\caption{Budget caps of the baseline workflows (the DAGent caps are listed in Table~\ref{tab:impl_framework}).}
\label{tab:impl_caps}
\centering
\small
\begin{tabular}{@{}lll@{}}
\toprule
\textbf{Workflow} & \textbf{Setting} & \textbf{Cap} \\
\midrule
ReAct Agent                & Context window                    & $32$K or $109$K tokens \\
Summary Agent / Fold Agent & Sessions per task                 & $10$ \\
\addlinespace
\multirow{2}{*}{Flash-Searcher} & Action steps per task        & $40$ \\
 & Summary interval                                            & $4$ (release default $8$) \\
\addlinespace
\multirow{6}{*}{FlowSearch} & Main iterations per task         & $5$ \\
 & Planner iterations / nodes                                  & $2$ / $7$ \\
 & Parallel workers                                            & $10$ \\
 & Sub-tasks / tool calls per node                             & $2$ / $5$ \\
 & Correction attempts                                         & $3$ \\
 & Task timeout                                                & $2{,}400$\,s \\
\addlinespace
\multirow{2}{*}{Plan-then-Patch variant} & Patch turns per task & $30$ \\
 & Session timeout                                             & $5{,}400$\,s \\
\bottomrule
\end{tabular}
\end{table}

\subsection{RL Training Hyperparameters}
\label{appendix:rl_hparams}

Table~\ref{tab:impl_rl} lists the RL training and DAGRPO hyperparameters. Training uses LoRA on all linear layers with asynchronous vLLM rollouts; unlisted settings inherit the VeRL defaults.

\begin{table}[!ht]
\caption{RL training and DAGRPO hyperparameters.}
\label{tab:impl_rl}
\centering
\small
\begin{tabular}{@{}ll@{}}
\toprule
\textbf{Parameter} & \textbf{Value} \\
\midrule
Hardware                                       & $2 \times$ H200, tensor parallel $2$ \\
LoRA rank / alpha                              & $64$ / $64$, all linear layers \\
Optimizer, learning rate                       & AdamW, $1 \times 10^{-5}$ \\
Tasks per batch, rollouts per task $G$         & $32$, $8$ \\
PPO mini-batch (tasks / trajectories)          & $16$ / $128$ \\
Asymmetric clip $\epsilon_{\ell}$ / $\epsilon_h$ & $0.20$ / $0.28$ \\
Entropy coefficient                            & $0.001$ \\
KL-to-reference coefficient                    & $0.001$ \\
Update steps                                   & $21$ \\
$\alpha$, DAGRPO / GRPO baseline               & $0.5$ / $1.0$ \\
$\lambda_{\text{proc}}$, $K_{\text{proc}}$     & $0.3$, $5$ \\
Training seeds                                 & $42$, $123$, $777$ \\
\bottomrule
\end{tabular}
\end{table}

\subsection{Human Agreement Study on the LLM Judge}
\label{appendix:human_agreement}

The Pass@1 numbers in \S\ref{sec:experiments} are produced by GPT-4o-mini with GPT-4.1 as a tie-breaker on borderline cases, applied through three dataset-specific judge prompts: an HLE-derived rubric for BrowseComp-Plus following its protocol \citep{chen2025browsecomp}, a short equivalence prompt for GAIA that replaces the original quasi-exact-match scoring of the benchmark \citep{mialon2024gaia} for cross-benchmark consistency, and a Chinese-language rubric for xbench-DeepSearch \citep{chen2025xbench}. None has been validated against humans in prior work, so we calibrate each judge against three annotators following \citet{du2025deepresearchbench}, \citet{wong2025widesearch}, and the LLM-as-judge meta-evaluation literature \citep{zheng2023judging, chan2024chateval}.

\paragraph{Setup.}
We sample $150$ (question, predicted answer, reference answer) tuples stratified $50/50/50$ across the three benchmarks, drawn proportionally from DAGent and four agent baselines so that correct and incorrect predictions appear in roughly equal share. Three volunteer annotators with reading-comprehension and information-retrieval backgrounds independently provide a forced binary judgment, blinded to the score of the LLM judge.

\obs{8}{Each per-benchmark Cohen's $\kappa$ between the human-majority label and the LLM judge falls in the almost-perfect bracket, with the aggregate $\kappa = 0.95$ approaching the human-noise ceiling.}
Table~\ref{tab:human_agreement} reports Fleiss' $\kappa$ on the three-human ensemble, Cohen's $\kappa$ between the human-majority label and the LLM judge, and raw percent agreement. All per-benchmark Cohen's $\kappa$ values fall in the almost-perfect bracket of \citet{landis1977agreement}; the aggregate $\kappa = 0.95$ exceeds the $\kappa = 0.19$ pairwise alignment of \citet{chan2024chateval} on FairEval, and the aggregate raw agreement of $97.3\%$ approaches the $97.8$ to $98.3\%$ range reported by \citet{wong2025widesearch} across four GPT-4.1-class judges, the small remaining gap reflecting our weaker GPT-4o-mini primary judge. Agreement is uniformly high on the English benchmarks and slightly lower on xbench-DeepSearch, where Chinese-language tasks admit longer phrasal answers and surface-form variation.

\begin{table}[!ht]
\caption{Human agreement on the LLM judge across $150$ (question, predicted answer, reference answer) tuples. Fleiss' $\kappa$ is computed on the three-human ensemble; Cohen's $\kappa$ is between the human-majority label and the LLM judge; raw \% is between the same pair.}
\label{tab:human_agreement}
\centering
\small
\begin{tabular}{@{}lcccc@{}}
\toprule
\textbf{Subset} & \textbf{N} & \textbf{Fleiss' $\kappa$ (3H)} & \textbf{Cohen's $\kappa$ (H vs J)} & \textbf{Raw \%} \\
\midrule
BrowseComp-Plus    & 50  & 0.98 & 0.96 & 98.0 \\
GAIA               & 50  & 0.98 & 0.96 & 98.0 \\
xbench-DeepSearch  & 50  & 0.96 & 0.92 & 96.0 \\
\midrule
Aggregate          & 150 & 0.97 & 0.95 & 97.3 \\
\bottomrule
\end{tabular}
\end{table}

\section{Additional Training-Free Results}
\label{appendix:tf_results}
This appendix collects the training-free results that supplement \S\ref{sec:main_results}, \S\ref{sec:ablation}, and \S\ref{sec:efficiency}.
\subsection{Cross-Backbone and Frontier-Scale Details}
\label{appendix:cross_backbone}

This section supplements the cross-backbone consistency check of \S\ref{sec:main_results} (Figure~\ref{fig:cross_backbone}) with the per-baseline breakdown in Table~\ref{tab:backbones}, plus the baseline list and multi-modal extension behind the frontier-scale comparison in Figure~\ref{fig:sota}. All rows share the tool stack of the main results, and the Qwen3-32B block of Table~\ref{tab:backbones} reproduces the corresponding rows of Table~\ref{tab:main}; FlowSearch, evaluated at Qwen3-32B only, is not part of this spot check.

\paragraph{Frontier-scale baselines.} For the GPT-5 327K comparison in Figure~\ref{fig:sota}, the BrowseComp-Plus baselines are 100B+ LLMs running ReAct at 327K context, namely Qwen3-235B-A22B, GLM-4.5-Air \citep{zeng2025glm45}, DeepSeek-V3.1 \citep{deepseek2025v31}, GPT-4.1 \citep{openai2025gpt41}, and GPT-5 \citep{openai2025gpt5}, with numbers reproduced from \citet{sun2025foldagent}; the GAIA and xbench-DeepSearch baselines are agent frameworks discussed in \S\ref{sec:rw_agents}, with numbers from \citet{qin2026flashsearcher} except OpenAI Deep Research and FlowSearch on GAIA, whose numbers are from \citet{hu2025flowsearch}. The GAIA runners-up Flash-Searcher (GPT-5), FlowSearch (o4-mini), and Skywork DR fall within a narrow range, between $82.4$ and $83.0$.

\paragraph{Multi-modal extension for the GAIA full validation set.} The GPT-5 327K comparison in Figure~\ref{fig:sota} evaluates GAIA on its full $165$-task validation set rather than the $103$-task text-only subset used elsewhere, following the protocol of \citet{qin2026flashsearcher}. The $62$ multi-modal questions in the full set involve attached image, audio, and document files that the text-only Orchestrator and ReAct Executor cannot process directly. Following the framework of \citet{zhu2025oagents}, we extend the tool registry with three inspector tools: an image inspector that base64-encodes attachments and queries a vision model, a document inspector that parses PDF, spreadsheet, and structured-text files for question-conditioned extraction, and an audio inspector that transcribes audio attachments before extraction. Each attached file is pre-processed into a textual description via the appropriate inspector and prepended to the question; the same inspectors remain available as agent-callable tools during execution. All other DAGent components remain unchanged, so the full-set evaluation isolates the multi-modal extension from the planning and context-management mechanisms of the framework.

\begin{table}[!ht]
\caption{Cross-backbone spot check with four mid-size open-source backbones; all training-free with the same tool stack as the main results; Pass@1 (\%); best per backbone in \textbf{bold}. The Qwen3-32B block reproduces the corresponding rows of Table~\ref{tab:main}.}
\label{tab:backbones}
\centering
\scriptsize
\setlength{\tabcolsep}{3pt}
\resizebox{\linewidth}{!}{%
\begin{tabular}{llllcccccccc}
\toprule
\multirow{2}{*}{\textbf{Backbone}} & \multirow{2}{*}{\textbf{Vendor}} & \multirow{2}{*}{\textbf{Agent Paradigm}} & \multicolumn{4}{c}{\textbf{BrowseComp-Plus}} & \multicolumn{4}{c}{\textbf{GAIA}} & \textbf{xbench-DS} \\
\cmidrule(lr){4-7} \cmidrule(lr){8-11}
 & & & Easy & Med. & Hard & Avg. & L1 & L2 & L3 & Avg. & Avg. \\
\midrule
\multirow{6}{*}{Qwen3-32B} & \multirow{6}{*}{Alibaba}
 & ReAct Agent (32K)        & 50.0 & 16.0 & 2.0 & 22.7 & 46.2 & 28.8 & 8.3 & 33.0 & 54.0 \\
 & & ReAct Agent (109K)     & 66.0 & 24.0 & 4.0 & 31.3 & 51.3 & 36.5 & 16.7 & 39.8 & 58.0 \\
 & & Summary Agent          & 80.0 & 36.0 & 0.0 & 38.7 & 51.3 & 28.8 & 25.0 & 36.9 & 55.0 \\
 & & Fold Agent             & 82.0 & 40.0 & 2.0 & 41.3 & 56.4 & 38.5 & 16.7 & 42.7 & 55.0 \\
 & & Flash-Searcher         & 80.0 & 34.0 & 2.0 & 38.7 & 56.4 & 40.4 & 25.0 & 44.7 & 58.0 \\
 & & \textbf{DAGent (Ours)} & \textbf{88.0} & \textbf{46.0} & \textbf{8.0} & \textbf{47.3} & \textbf{69.2} & \textbf{50.0} & \textbf{33.3} & \textbf{55.3} & \textbf{65.0} \\
\midrule
\multirow{6}{*}{Seed-OSS-36B} & \multirow{6}{*}{ByteDance}
 & ReAct Agent (32K)        & 64.0 & 18.0 & 0.0 & 27.3 & 43.6 & 40.4 & 8.3 & 37.9 & 58.0 \\
 & & ReAct Agent (109K)     & 88.0 & 44.0 & 2.0 & 44.7 & 43.6 & 48.1 & 16.7 & 42.7 & 62.0 \\
 & & Summary Agent          & 82.0 & 40.0 & 2.0 & 41.3 & 48.7 & 51.9 & \textbf{25.0} & 47.6 & 59.0 \\
 & & Fold Agent             & 88.0 & 36.0 & 8.0 & 44.0 & 43.6 & 46.2 & 16.7 & 41.7 & 60.0 \\
 & & Flash-Searcher         & 88.0 & 44.0 & 4.0 & 45.3 & 51.3 & 51.9 & \textbf{25.0} & 48.5 & 67.0 \\
 & & \textbf{DAGent (Ours)} & \textbf{94.0} & \textbf{58.0} & \textbf{10.0} & \textbf{54.0} & \textbf{64.1} & \textbf{53.8} & \textbf{25.0} & \textbf{54.4} & \textbf{69.0} \\
\midrule
\multirow{6}{*}{GLM-4-32B} & \multirow{6}{*}{Zhipu AI}
 & ReAct Agent (32K)        & 52.0 & 16.0 & 2.0 & 23.3 & 41.0 & 25.0 & 25.0 & 31.1 & 56.0 \\
 & & ReAct Agent (109K)     & 68.0 & 24.0 & 4.0 & 32.0 & 48.7 & 32.7 & 25.0 & 37.9 & 60.0 \\
 & & Summary Agent          & 82.0 & 36.0 & 2.0 & 40.0 & 43.6 & 30.8 & 25.0 & 35.0 & 57.0 \\
 & & Fold Agent             & 84.0 & 40.0 & 4.0 & 42.7 & 51.3 & 36.5 & 25.0 & 40.8 & 58.0 \\
 & & Flash-Searcher         & 82.0 & 34.0 & 2.0 & 39.3 & 53.8 & 38.5 & 25.0 & 42.7 & 64.0 \\
 & & \textbf{DAGent (Ours)} & \textbf{90.0} & \textbf{48.0} & \textbf{8.0} & \textbf{48.7} & \textbf{66.7} & \textbf{46.2} & \textbf{33.3} & \textbf{52.4} & \textbf{67.0} \\
\midrule
\multirow{6}{*}{Nemotron-3-Nano-30B} & \multirow{6}{*}{NVIDIA}
 & ReAct Agent (32K)        & 50.0 & 18.0 & 4.0 & 24.0 & 43.6 & 25.0 & 25.0 & 32.0 & 46.0 \\
 & & ReAct Agent (109K)     & 66.0 & 26.0 & 8.0 & 33.3 & 51.3 & 36.5 & 25.0 & 40.8 & 51.0 \\
 & & Summary Agent          & 80.0 & 38.0 & 2.0 & 40.0 & 53.8 & 30.8 & 25.0 & 38.8 & 48.0 \\
 & & Fold Agent             & 84.0 & 40.0 & 4.0 & 42.7 & 56.4 & 38.5 & 25.0 & 43.7 & 49.0 \\
 & & Flash-Searcher         & 82.0 & 38.0 & 4.0 & 41.3 & 59.0 & 40.4 & 25.0 & 45.6 & 53.0 \\
 & & \textbf{DAGent (Ours)} & \textbf{90.0} & \textbf{48.0} & \textbf{12.0} & \textbf{50.0} & \textbf{71.8} & \textbf{51.9} & \textbf{33.3} & \textbf{57.3} & \textbf{59.0} \\
\bottomrule
\end{tabular}%
}
\end{table}

\subsection{Plan-then-Patch Variant Details}
\label{appendix:ptp_details}

The Plan-then-Patch variant used in Table~\ref{tab:ablation} and Figure~\ref{fig:efficiency}(b) reuses the Orchestrator, ReAct Executor, backbone, tool stack, context-management components, and per-node refine budget of DAGent; only the cross-iteration planning behavior of the Orchestrator is modified.

\paragraph{Planning and patching protocol.}
At iteration $0$ the Orchestrator outputs the complete task DAG in one call, committing every search node plus exactly one answer node with full id, description, prompt, and dependency wiring before any node has executed; subsequent iterations execute the next ready batch in parallel and defer the answer node whenever any non-answer node is also ready. After each batch the Orchestrator may issue \emph{Refine} operations, reusing the three-attempt mechanism of DAGent on a node whose status is Uncertain or Not Found, or \emph{Add} operations, introducing a new search node whose description must cite at least one already-executed node id (enforced by the structural validator through the format-warning loop). Both operations preserve the DAGent JSON schema; added nodes are automatically appended to the dependency set of the answer node so their evidence reaches final synthesis. Total patch turns are capped at $30$, with mean per-task patch turns of $4.2 / 1.8 / 1.4$ on BrowseComp-Plus / GAIA / xbench-DeepSearch.

\paragraph{Absolute per-task footprints.}
Table~\ref{tab:efficiency_extended} in Appendix~\ref{appendix:efficiency_extended} lists the absolute per-task footprints summarized in Figure~\ref{fig:efficiency}(b), alongside those of Flash-Searcher and FlowSearch.

\paragraph{Where the off-chain nodes come from.}
Figure~\ref{fig:efficiency}(b) shows that the variant raises the off-chain Executor ratio from 0.25 / 0.20 / 0.18 to 0.40 / 0.30 / 0.25. Where these off-chain nodes come from follows from how the variant is built rather than from a measurement: every patch node, whether a refine or an add, is connected to the answer node automatically so that its evidence reaches the final synthesis, so patch nodes are on-chain by construction and off-chain nodes can only come from the initial plan. The extra off-chain nodes are therefore over-commitment at planning time, not a failure of the patch mechanism to prune. A smarter deletion policy would close only part of the gap: deleting a node saves compute only if the node has not yet run, and a patch policy that decides batch by batch, from evidence, what to keep, drop, or expand is Evaluate-then-Grow. The released FlowSearch implementation includes a deletion-capable refiner, and Table~\ref{tab:efficiency_extended} shows that it still costs more than DAGent on every cost column.

\subsection{Extended Efficiency Comparison}
\label{appendix:efficiency_extended}

Section~\ref{sec:efficiency} compares DAGent with its Plan-then-Patch variant. This section extends the comparison to the two external Plan-then-Patch systems, Flash-Searcher and FlowSearch, and reports absolute values for external tool calls and wall-clock time, which the main text quotes only as relative differences.

\paragraph{Setting.}
All four workflows use the same Qwen3-32B backbone and the same retrieval backends on every benchmark, and no method is modified beyond the configuration documented in Appendix~\ref{appendix:baseline_caps}; FlowSearch runs with its Coordinator enabled. The \emph{tool calls} column counts every model-driven call in the workflow, so its composition differs by method: search, open\_page, and recall, plus Orchestrator plan calls for the DAGent rows, planning and summary calls for Flash-Searcher, and Planner and Coordinator calls for FlowSearch. The \emph{external tool calls} column counts only search and open\_page, which all four workflows use in the same way, and therefore isolates the cost of the search API. The \emph{steps} column follows the parallelism-aware definition of \S\ref{sec:efficiency} and does not measure latency. The \emph{tokens} column is total input plus output, and the \emph{time} column is the wall-clock time to finish one task. The \emph{nodes} column is the final graph size and \emph{off-chain} is the fraction of Executor nodes outside the answer-inclusive closure; both are defined only for the two DAGent rows. The \emph{calls / step} column is the ratio of tool calls to execution steps and is reported as a parallelism measure, not as a cost. We do not impose a common per-task cost cap on the four workflows, because most workflows stop through an explicit finish call rather than by exhausting a budget (Appendix~\ref{appendix:baseline_caps} lists the caps of each workflow), and a common cap would cut the remaining runs off mid-task and turn the comparison into a completion-rate test.

\begin{table}[!ht]
\caption{Per-task cost of four workflows with Qwen3-32B and the same retrieval backends. Pass@1 values are from Table~\ref{tab:main} and Table~\ref{tab:ablation}. Tool calls, external tool calls, execution steps, tokens, and time are per-task means; the calls / step column is the sum of tool calls divided by the sum of execution steps.}
\label{tab:efficiency_extended}
\centering
\small
\setlength{\tabcolsep}{3.5pt}
\resizebox{\linewidth}{!}{%
\begin{tabular}{@{}llccccccccc@{}}
\toprule
\textbf{Benchmark} & \textbf{Workflow} & \textbf{Pass@1} & \textbf{Tool calls} & \textbf{External calls} & \textbf{Steps} & \textbf{Calls / step} & \textbf{Tokens (M)} & \textbf{Time (s)} & \textbf{Nodes} & \textbf{Off-chain} \\
\midrule
\multirow{4}{*}{BrowseComp-Plus}
 & DAGent (Full)            & \textbf{47.3} & 67.6 & 60.8 & 42.7 & 1.58 & 1.20 & 541.6 & 11.8 & 0.25 \\
 & Plan-then-Patch variant  & 42.0 & 87.9 & 82.3 & 54.8 & 1.60 & 1.68 & 731.9 & 17.6 & 0.40 \\
 & Flash-Searcher           & 38.7 & 51.3 & 42.1 & 35.8 & 1.43 & 0.91 & 388.7 & -- & -- \\
 & FlowSearch               & 43.3 & 83.5 & 76.7 & 47.6 & 1.75 & 1.57 & 714.5 & -- & -- \\
\midrule
\multirow{4}{*}{GAIA}
 & DAGent (Full)            & \textbf{55.3} & 36.8 & 33.1 & 31.2 & 1.18 & 0.66 & 461.8 & 5.4 & 0.20 \\
 & Plan-then-Patch variant  & 50.5 & 45.3 & 42.0 & 37.6 & 1.20 & 0.85 & 584.6 & 7.3 & 0.30 \\
 & Flash-Searcher           & 44.7 & 28.7 & 21.8 & 23.5 & 1.22 & 0.48 & 301.9 & -- & -- \\
 & FlowSearch               & 51.5 & 40.4 & 35.1 & 32.8 & 1.23 & 0.74 & 567.7 & -- & -- \\
\midrule
\multirow{4}{*}{xbench-DeepSearch}
 & DAGent (Full)            & \textbf{65.0} & 32.9 & 29.2 & 26.8 & 1.23 & 0.44 & 380.9 & 4.9 & 0.18 \\
 & Plan-then-Patch variant  & 61.0 & 38.2 & 35.4 & 32.4 & 1.18 & 0.52 & 466.8 & 6.2 & 0.25 \\
 & Flash-Searcher           & 58.0 & 24.6 & 18.9 & 20.2 & 1.22 & 0.30 & 241.6 & -- & -- \\
 & FlowSearch               & 63.0 & 36.7 & 31.6 & 29.6 & 1.24 & 0.55 & 488.9 & -- & -- \\
\bottomrule
\end{tabular}%
}
\end{table}

\obs{9}{Among the systems that keep a separate context per node, DAGent is both the most accurate and the cheapest on every per-task cost column (tool calls, external calls, steps, tokens, and time); Flash-Searcher spends less because it is a single-agent system, and it is 8.6 / 10.6 / 7.0 points less accurate.}
The Plan-then-Patch variant differs from DAGent only in when it commits to a plan, and it makes 35 / 27 / 21\% more external tool calls (82.3 / 42.0 / 35.4 against 60.8 / 33.1 / 29.2) and needs 35 / 27 / 23\% more wall-clock time (731.9 / 584.6 / 466.8\,s against 541.6 / 461.8 / 380.9\,s). FlowSearch makes 26 / 6 / 8\% more external tool calls and needs 32 / 23 / 28\% more time, while staying 4.0 / 3.8 / 2.0 points below DAGent. A pre-committed plan contains branches that later turn out not to be needed, and the system still has to execute them; when the graph grows only after evidence arrives, those branches are never created. Executing them is what accounts for the extra search calls and wall-clock time of the two pre-committed workflows. Flash-Searcher is the one workflow that costs less than DAGent, and the reason is architectural rather than a property of Plan-then-Patch: it keeps a single reasoning trajectory and merges every parallel branch back into the same state, so it never holds a separate context per node. DAGent, its Plan-then-Patch variant, and FlowSearch are multi-agent systems that keep one context per node, which accounts for the additional tokens and calls and is also what lets a DAG operate over long horizons. Lower cost only matters at comparable accuracy, and no workflow in Table~\ref{tab:efficiency_extended} matches the accuracy of DAGent at any cost level.

\paragraph{Overall accuracy of the Plan-then-Patch systems.}
Averaged over the three benchmarks, FlowSearch reaches 52.6, the Plan-then-Patch variant 51.2, and Flash-Searcher 47.1, against 55.9 for DAGent. The variant, which keeps the full DAGent component stack, therefore scores between the two released Plan-then-Patch systems rather than below both, so Table~\ref{tab:ablation} does not compare Evaluate-then-Grow against a weakened Plan-then-Patch system, and the released FlowSearch implementation, whose refiner can also delete nodes, is the strongest non-DAGent system in the Qwen3-32B block of Table~\ref{tab:main} while still costing more than DAGent on every cost column.

\subsection{Behavioral Statistics of the Evaluate-then-Grow Loop}
\label{appendix:behavior}

\begin{figure}[!ht]
\centering
\includegraphics[width=\linewidth]{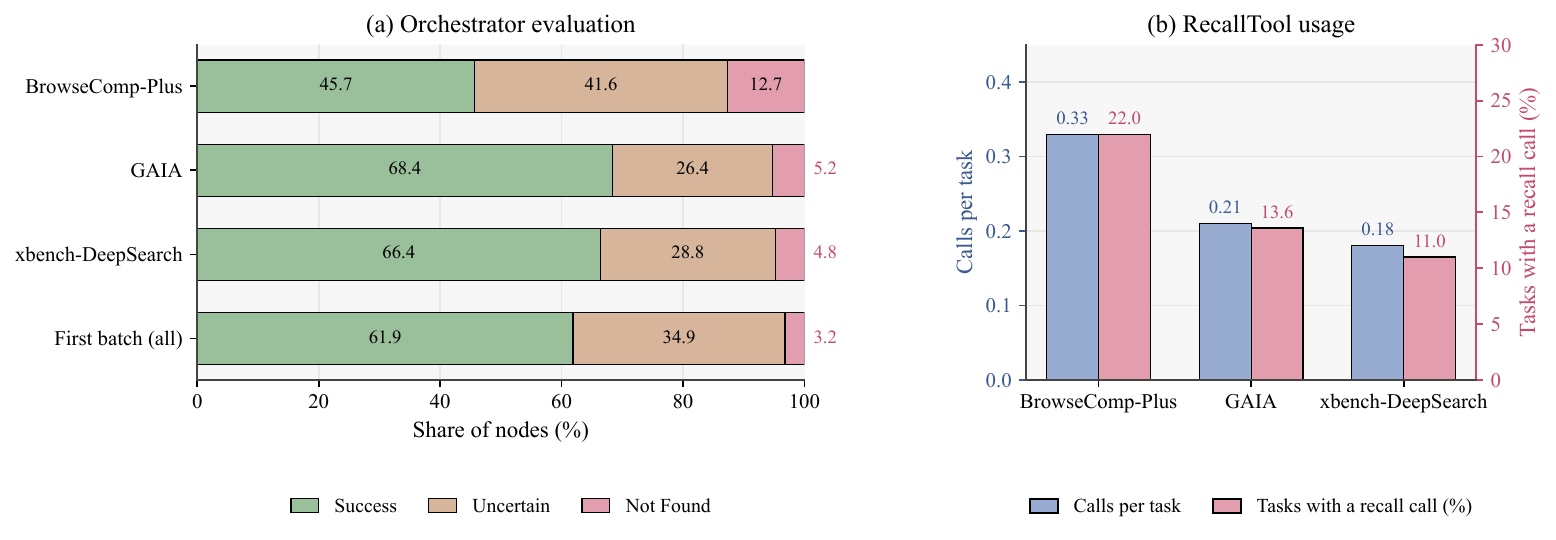}
\caption{Behavioral statistics of the Qwen3-32B training-free runs. (a) Distribution of the Orchestrator evaluation outcomes over all executed nodes per benchmark and over first-batch nodes pooled across the three benchmarks. (b) \textsc{RecallTool} usage: mean number of calls per task and share of tasks that make at least one call, per benchmark.}
\label{fig:behavior}
\end{figure}

Figure~\ref{fig:behavior} summarizes statistics computed from the logs of the Qwen3-32B training-free runs reported in the main text: how the statuses that the Orchestrator evaluation assigns are distributed over all executed nodes and over first-batch nodes, and how often the Executor calls \textsc{RecallTool}.

\obs{10}{Inconclusive evaluations are a routine planning input rather than an exception, and recall stays a fallback: \emph{Uncertain} and \emph{Not Found} account for 54.3\% of the executed nodes on BrowseComp-Plus and about a third on GAIA and xbench-DeepSearch, while 22.0 / 13.6 / 11.0\% of tasks make at least one recall call.}

\paragraph{\textsc{RecallTool} usage.}
\label{appendix:recall_usage}
The Executor calls \textsc{RecallTool} (Eq.~\eqref{eq:recall}) as \texttt{recall(node\_id, goal)}. The extractor reads the full InteractionTranscript of the named dependency and returns what the goal asks for; it may use only that transcript, must attach a document id, URL, or exact quote to every claim, and must state explicitly when nothing relevant is present (prompt in Appendix~\ref{appendix:prompts}). The tool is a fallback rather than a default channel: the mean number of calls per task is 0.33 / 0.21 / 0.18 on BrowseComp-Plus / GAIA / xbench-DeepSearch, and 22.0\% / 13.6\% / 11.0\% of tasks make at least one call (Figure~\ref{fig:behavior}b). The w/o InteractionTranscript row of Table~\ref{tab:ablation} is the corresponding ablation, since removing the transcript disables recall, and it costs 4.0 / 2.9 / 2.0 Pass@1 points; given the usage rates, the tool changes the outcome of roughly one in five of the tasks that call it.

\paragraph{Evaluation outcomes and the first batch.}
\label{appendix:first_batch}
Across all executed nodes, the Orchestrator evaluates 41.6\% / 12.7\% of nodes as \emph{Uncertain} / \emph{Not Found} on BrowseComp-Plus, 26.4\% / 5.2\% on GAIA, and 28.8\% / 4.8\% on xbench-DeepSearch (Figure~\ref{fig:behavior}a). The Orchestrator reads these outcomes to decide the next expansion, so negative and inconclusive evidence is a routine planning input rather than discarded work. First-batch nodes are planned from the task description alone (Appendix~\ref{appendix:rationale_planning}) and are evaluated as \emph{Success} / \emph{Uncertain} / \emph{Not Found} in 61.9\% / 34.9\% / 3.2\% of cases, so a first node that finds nothing relevant at all is rare. The validator forces a refine node for every weak first node in the same turn; these refinements turn 44.4\% of weak first nodes into \emph{Success}, and the remainder reach the next expansion as an explicit uncertainty rather than as silently accepted evidence.

\paragraph{Structural violations and accuracy.}
\label{appendix:violations}
The structural compliance regularization of DAGRPO penalizes Orchestrator turns that violate the constraints of \S\ref{sec:dagrpo}. Removing the term costs 1.6 / 1.0 / 0.7 Pass@1 points (Table~\ref{tab:dagrpo_analysis}), and Appendix~\ref{appendix:dagrpo_dynamics} shows that it drives the format-warning rate of the Orchestrator down faster than the outcome-only baseline. The association is also visible in the evaluation logs: tasks whose trajectory contains at least one structural violation score 14.7 points lower than tasks with none (43.6 against 58.3). Task difficulty may contribute to both, so we read this as an association consistent with the mechanism rather than as a causal estimate.

\section{Additional RL Results}
\label{appendix:rl_results}
This appendix collects the RL results that supplement the training-based rows of Table~\ref{tab:main}.
\subsection{Per-Seed Training Results}
\label{appendix:per_seed}

\begin{figure}[!ht]
\centering
\includegraphics[width=\linewidth]{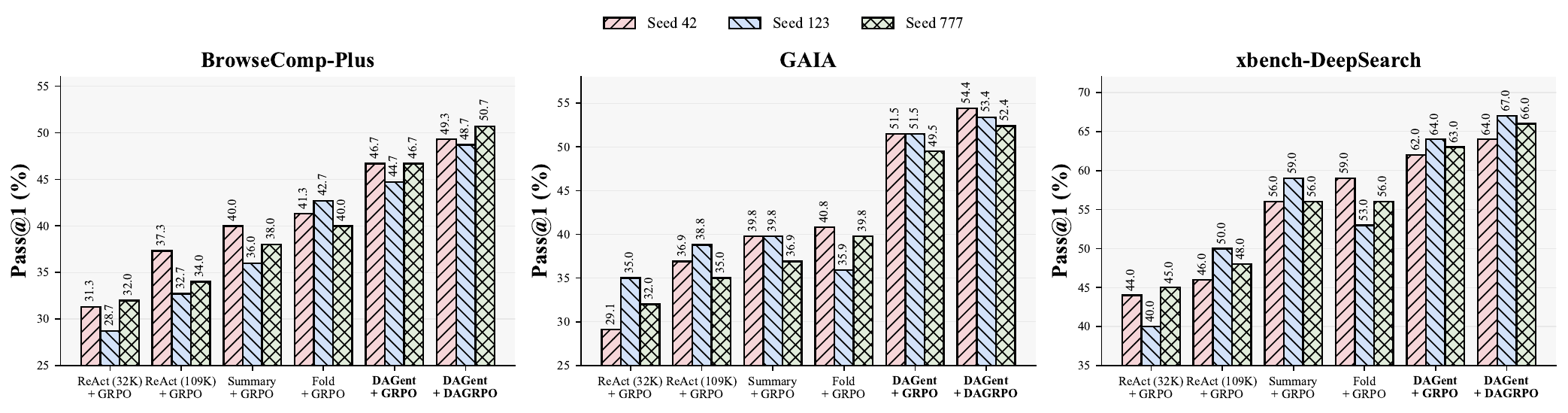}
\caption{Per-seed Pass@1 (\%) for the six RL-trained variants in the Qwen3-8B block of Table~\ref{tab:main}, broken out by training seed (42 / 123 / 777) on BrowseComp-Plus, GAIA, and xbench-DeepSearch.}
\label{fig:per_seed}
\end{figure}

Three independent training seeds ($42$, $123$, $777$) are run for each of the six RL-trained variants in the Qwen3-8B block of Table~\ref{tab:main} with all other hyperparameters held fixed; the seed enters through both the dataloader shuffle and the LoRA initialization. Figure~\ref{fig:per_seed} reports the per-seed Pass@1 (\%) values of these variants, one panel per benchmark.

\obs{11}{The RL gains of DAGent are robust across training seeds, with standard deviations below $1.6$ absolute points and no single seed reversing the DAGRPO-versus-GRPO ranking.}
Standard deviations of the two RL rows of DAGent in the Qwen3-8B block of Table~\ref{tab:main} stay below $1.6$ across all three benchmarks, and the per-seed values in Figure~\ref{fig:per_seed} show that no single seed alone reverses the DAGRPO-versus-GRPO ranking on any benchmark.

\subsection{DAGRPO Training Dynamics}
\label{appendix:dagrpo_dynamics}

\begin{figure}[!ht]
\centering
\includegraphics[width=\linewidth]{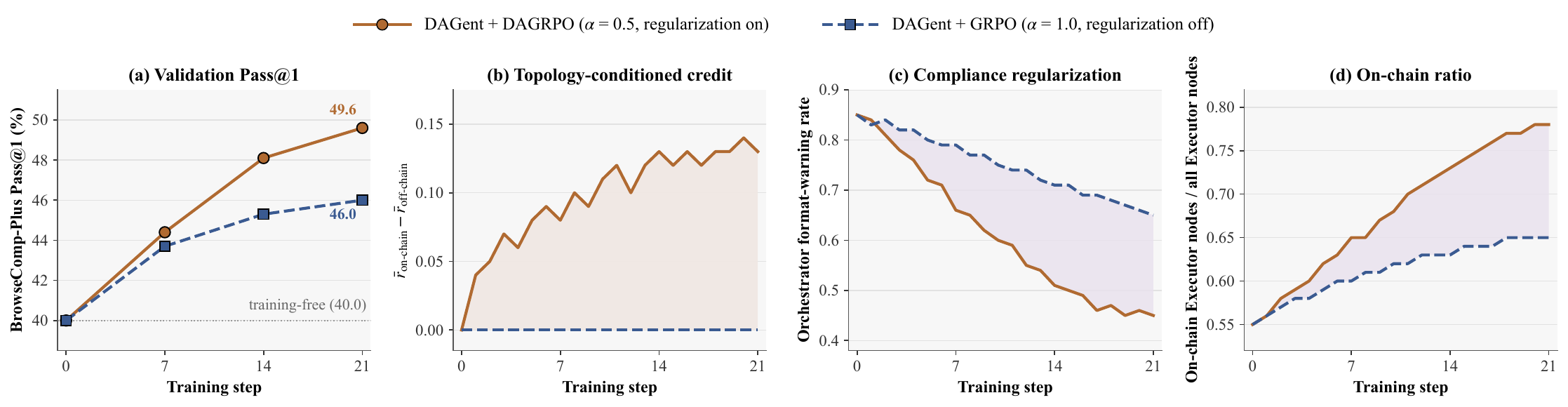}
\caption{DAGRPO training dynamics on Qwen3-8B with LoRA. (a) BrowseComp-Plus validation Pass@1 over training steps, averaged over the three training seeds, so that the end points equal the Qwen3-8B means of Table~\ref{tab:main}. (b) Per-step training-time reward gap $\bar{r}_{\mathrm{chain}} - \bar{r}_{\mathrm{off\text{-}chain}}$; the $\alpha = 0.5$ topology credit separates the two groups, while $\alpha = 1.0$ gives zero gap by construction. (c) Per-step Orchestrator format-warning rate from the structural validator. (d) Per-step on-chain ratio, the fraction of Executor nodes inside the answer-inclusive closure.}
\label{fig:dagrpo_dynamics}
\end{figure}

Figure~\ref{fig:dagrpo_dynamics} traces four training-time metrics that distinguish DAGRPO from GRPO; panels (b) to (d) are computed on the sampled training rollouts of each step, so their levels differ from the evaluation-time values of Table~\ref{tab:rl_efficiency}, which are measured on the evaluation set with greedy decoding.

\obs{12}{Both DAGRPO signals act as designed during training: the topology-conditioned credit separates the on-chain and off-chain Executor rewards by construction, which the $\alpha = 1.0$ baseline never does, and the compliance regularization lowers the Orchestrator format-warning rate faster than outcome-only GRPO.}
Panel (b) shows the topology credit at $\alpha = 0.5$ separating on-chain and off-chain Executor reward by construction, while the $\alpha = 1.0$ baseline shows zero gap. Panel (c) shows the compliance regularization driving the Orchestrator format-warning rate down faster than the outcome-only baseline. Panels (a) and (d) report validation Pass@1 and on-chain-ratio progression over training, included for completeness of the training record.

\subsection{Efficiency as a Byproduct of RL Training}
\label{appendix:rl_efficiency}

We did not design DAGRPO for efficiency: neither the topology-conditioned credit nor the structural compliance regularization targets per-task footprint. Yet across the three benchmarks we observe consistent reductions on every footprint metric, reported here as a byproduct of accuracy-driven RL training averaged over three independent training seeds.

\obs{13}{RL training reduces the per-task footprint of DAGent without any explicit efficiency design, and DAGRPO reduces it further than role-separated GRPO.}
Role-separated outcome-only GRPO (the GRPO-DAGent row of Table~\ref{tab:main}) yields step reductions of $8$ to $12\%$, tool-call reductions of $6$ to $9\%$, and token reductions of $15$ to $16\%$ across BrowseComp-Plus / GAIA / xbench-DeepSearch (Table~\ref{tab:rl_efficiency}); the two structural signals of DAGRPO raise these to $12$ to $17\%$, $10$ to $14\%$, and $17$ to $21\%$, in line with reductions reported across step-grouped \citep{feng2025gigpo} and DAG-structured \citep{lu2025orchdag, wu2025gap} RL agent variants. Structural metrics decrease in parallel: the graph node count drops by $0.7$ to $0.9$ nodes per task under GRPO and by $1.3$ to $1.9$ under DAGRPO, and the off-chain Executor ratio decreases by $0.02$ to $0.04$ under GRPO and by $0.05$ to $0.09$ under DAGRPO. The additional reduction under DAGRPO is spread evenly over steps, tool calls, and tokens, between 2 and 5 points of the training-free value each, and is largest for the off-chain ratio, consistent with the pressure of the topology-conditioned credit on answer-chain focus.

\begin{table}[!ht]
\caption{Effect of RL training on the per-task efficiency of DAGent on Qwen3-8B; the RL rows are means across three independent training seeds. Pass@1 values are taken from the Qwen3-8B block of Table~\ref{tab:main}. Tokens is total input plus output; tool calls aggregate search, open\_page, recall, and Orchestrator plan invocations; nodes is the final graph size; off-chain is the fraction of Executor nodes outside the answer-inclusive closure, as in Table~\ref{tab:efficiency_extended}.}
\label{tab:rl_efficiency}
\centering
\small
\setlength{\tabcolsep}{4pt}
\begin{tabular}{llcccccc}
\toprule
\textbf{Benchmark} & \textbf{Setting} & \textbf{Pass@1} & \textbf{Tokens} & \textbf{Tool calls} & \textbf{Steps} & \textbf{Nodes} & \textbf{Off-chain} \\
\midrule
\multirow{3}{*}{BrowseComp-Plus}
& DAGent (training-free)         & 40.0 & 1.14M & 65.5 & 41.5 & 12.2 & 0.25 \\
& \quad + RL (GRPO)             & 46.0 & 0.96M & 59.6 & 36.5 & 11.3 & 0.21 \\
& \quad + RL (DAGRPO)           & 49.6 & 0.90M & 56.3 & 34.5 & 10.3 & 0.16 \\
\midrule
\multirow{3}{*}{GAIA}
& DAGent (training-free)         & 46.6 & 0.62M & 35.5 & 29.5 & 5.3 & 0.20 \\
& \quad + RL (GRPO)             & 50.8 & 0.53M & 33.0 & 26.8 & 4.6 & 0.17 \\
& \quad + RL (DAGRPO)           & 53.4 & 0.51M & 31.6 & 25.7 & 4.0 & 0.13 \\
\midrule
\multirow{3}{*}{xbench-DeepSearch}
& DAGent (training-free)         & 60.0 & 0.41M & 32.0 & 26.0 & 5.2 & 0.18 \\
& \quad + RL (GRPO)             & 63.0 & 0.35M & 30.1 & 23.9 & 4.5 & 0.16 \\
& \quad + RL (DAGRPO)           & 65.7 & 0.34M & 28.8 & 22.9 & 3.9 & 0.13 \\
\bottomrule
\end{tabular}
\end{table}

\section{Limitations}
\label{appendix:limitations}

DAGent is text-native. The multi-modal extension used for the full GAIA validation set (Appendix~\ref{appendix:cross_backbone}) converts attachments into text before planning begins; handling multi-modal evidence inside the loop would require changes to how the Orchestrator evaluates evidence and how the Executor represents its transcript, not only additional tools. DAGent also plans over the evidence that its tools return and cannot create evidence that the retriever never retrieves, so its accuracy depends on search-engine, retrieval-embedding, and page-extraction quality. The answer-inclusive closure is a structural proxy for contribution rather than a causal attribution, and the association between structural compliance and accuracy that we report is correlational (Appendices~\ref{sec:rationale} and~\ref{appendix:behavior}); per-node causal attribution in DAG-based multi-agent systems remains open. The three benchmarks have closed-form answers that an LLM judge can verify, so open-ended report generation is outside the evidence presented here. Finally, the RL experiments train one backbone (Qwen3-8B) with LoRA for 21 update steps over three seeds; whether the DAGRPO gains persist at larger training scales, longer schedules, or full fine-tuning is not established by this paper.

\section{Prompts}
\label{appendix:prompts}

This appendix lists, verbatim, the prompts of the three DAGent roles and the three LLM judge prompts. Inside a listing, section headings are set in bold in the color of the role, and template slots that are filled at run time, such as \texttt{\{goal\}}, are set in orange. The Orchestrator prompt is shown with the Overall Goal of one BrowseComp-Plus task filled in.

\subsection{Orchestrator System Prompt}
\label{appendix:prompt_orchestrator}
\begin{promptbox}[Orchestrator System Prompt]{roleOrch}
\begin{vbPrompt}
colortcBodyYou are a world-class orchestrator for deep research. You plan search tasks for a ReAct agent,
colortcBodyevaluate its results, and iterate until the Overall Goal is answered.
colortcBody
colorroleOrchtextbf## Overall Goal
colortcBodyAn individual was encouraged to use a different tool for their work during the year between the
colortcBodyassassinations of John F. Kennedy and his brother, Senator Robert F. Kennedy. Up until 2014, they
colortcBodyhave continued to use this tool. This person first studied at a university in Africa, then in the
colortcBodyUnited Kingdom, and later in the United States. They taught and held the positions of head of
colortcBodydepartment and dean at the university where they studied in Africa until a year in the last decade
colortcBodyof the 20th century. They also built a venue to showcase their own work as well as the work of
colortcBodyother people. This venue celebrated its 25th anniversary in a year between 2015 and 2020. What is
colortcBodythe name of this individual?
colortcBody
colorroleOrchtextbf## How This System Works
colortcBodyYou manage nodes in a task graph. Your conversation history contains all previous planning turns.
colortcBodyEach new user message only contains NEW execution results and a compact status ledger. Review your
colortcBodyhistory to recall earlier node details.
colortcBody
colorroleOrchtextbf## Core Responsibilities
colortcBody
colorroleOrchtextbf### 1. Evaluate (output: `node_statuses`)
colortcBodyFor each node listed as "Awaiting Evaluation" in the user message, evaluate its execution result
colortcBodyand assign status:
colortcBody- textbf**Success**: information retrieved with high confidence, grounded in actual search results.
colortcBody- textbf**Uncertain**: information found but verification needed (triggers refine).
colortcBody- textbf**Not Found**: search executed correctly but yielded no results (triggers refine).
colortcBody
colortcBodyEvery id listed under "Nodes to evaluate this turn" MUST appear in `node_statuses`. Do NOT include
colortcBodynodes that already have a status.
colortcBody
colorroleOrchtextbf### 2. Plan (output: `nodes`)
colortcBodyPlan nodes that can be executed immediately. Each node has `id`, `description`, `prompt`,
colortcBody`dependencies`.
colortcBody
colortcBodytextbf**id** (encodes task type):
colortcBody- Search: `t1`, `t2`, `t3`, ...
colortcBody- Refine: `<parent>_refine1` or `<parent>_refine2` (id MUST end in `_refineN`)
colortcBody- Answer: `t_answer` (id MUST end in exactly `_answer`)
colortcBody
colortcBodytextbf**description**: must be detailed and textbf**self-contained**. Include all necessary context.
colortcBody
colortcBodytextbf**prompt**: provide detailed instructions for the ReAct agent. Synthesize relevant findings from
colortcBodydependency nodes as context bridge, and add operational value such as search strategies, angles to
colortcBodytry, and prior results to avoid. Formulate textbf**complete natural language sentences** (NOT keywords)
colortcBodyfor the ReAct agent. Our retrieval uses embedding model - full sentences work much better. Preserve
colortcBodyrare verbatim phrases from the Overall Goal exactly.
colortcBody
colortcBodytextbf**dependencies**: list ALL nodes whose content is needed. All dependencies must already be
colortcBodyevaluated (not Awaiting Evaluation, not in the same batch).
colortcBody
colortcBodytextbf**IMPORTANT**: the ReAct agent can ONLY see the description, prompt, and content from dependency
colortcBodynodes. It CANNOT see the Overall Goal or full graph state. Use these three fields together to give
colortcBodyit everything it needs.
colortcBody
colortcBodytextbf**Granularity Rule**: task granularity depends on the purpose:
colortcBody- textbf**Finding candidates**: first, resolve any conditions that contain indirect references
colortcBody(references to events, abstract categories that need resolution into specific entity names) into
colortcBodyconcrete facts via search. If a lookup returns multiple plausible values, try each value in
colortcBodyseparate candidate-discovery searches. Then, substitute resolved facts into the remaining
colortcBodyconditions and combine 2-3 distinctive conditions per node to narrow down candidates. Plan MULTIPLE
colortcBodycandidate-discovery nodes in the SAME turn with DIFFERENT condition combinations. Do NOT put all
colortcBodyconditions in one node (retrieval degrades with too many conditions). After gathering candidates,
colortcBodyproceed to check them.
colortcBody- textbf**Checking specific conditions**: each node should target ONE condition for ONE candidate.
colortcBody- textbf**Fallback**: if all candidates fail critical conditions, return to finding candidates with
colortcBodyDIFFERENT search angles. If candidate discovery keeps failing, the resolved facts themselves may be
colortcBodywrong. Re-do those lookups with different angles.
colortcBody
colortcBodytextbf**Parallelism Rule**: you can include MULTIPLE nodes per plan.
colortcBody- When finding candidates, plan MULTIPLE search nodes in parallel to explore DIFFERENT combinations.
colortcBody- When checking conditions for one candidate, plan ALL unchecked conditions in parallel.
colortcBody- Dependency constraint: parallel nodes can only be planned if all their dependencies have already
colortcBodybeen evaluated.
colortcBody
colortcBodytextbf**Refine Rule**:
colortcBody- If status is Uncertain or Not Found AND the line still has refine budget, you MUST plan a refine
colortcBodynode (`<parent>_refine1`, or `<parent>_refine2` for a second refine) with a DIFFERENT search
colortcBodystrategy. The refine prompt MUST copy ALL prior search queries from the entire refine chain
colortcBodyverbatim (the original node's `key_steps` plus any earlier refine's `key_steps`) and specify new
colortcBodyangles.
colortcBody- textbf**3-attempt limit**: original search + refine1 + refine2. After 3 attempts, the line is
colortcBodyexhausted. Accept that gap and move forward.
colortcBody
colortcBodytextbf**Answer Rule**:
colortcBody- Plan a single answer node when: (1) evidence is sufficient AND a concrete candidate exists, OR
colortcBody(2) all search lines are exhausted AND a best-effort answer is available. If a `[CONTEXT WARNING]`
colortcBodyhas been injected, begin the termination path: verify the answer's required form if needed, then
colortcBodyplan the answer node.
colortcBody- Finding candidates is NOT enough. You MUST go through the checking-conditions phase first.
colortcBody- You MUST have a concrete candidate. The question is guaranteed to have an answer. Responses like
colortcBody"No such X exists" or "Insufficient information" are WRONG. Compare candidates by how many
colortcBodyconditions each satisfies and pick the best-supported one.
colortcBody- Before the answer node, try to ensure all necessary information about the selected candidate has
colortcBodybeen gathered by upstream nodes. If the Overall Goal requires a specific form or granularity,
colortcBodyideally an upstream node should have already retrieved it. If not, the answer node can perform
colortcBodytargeted searches to fill in missing details.
colortcBody- The answer node MUST be planned ALONE, NEVER in the same batch with search or refine nodes. Set
colortcBodyits prompt to list accumulated evidence and instruct the ReAct agent to verify any details still
colortcBodyneeded for the answer's required form, then call `finish`.
colortcBody
colorroleOrchtextbf## Output Format
colortcBody
colortcBodyStructure your response in TWO parts.
colortcBody
colorroleOrchtextbf### Part 1: Thinking
colortcBodyYou MUST answer EACH question below step by step. Do NOT include any JSON in this section.
colortcBody
colortcBodytextbf**Step 1: Evaluate**
colortcBody- What status (Success / Uncertain / Not Found) should I assign to each Awaiting Evaluation node?
colortcBodyIs the answer grounded in actual search results, or guessed from prior knowledge?
colortcBody- Have I covered every id listed in "Nodes to evaluate this turn", and no others?
colortcBody
colortcBodytextbf**Step 2: Plan**
colortcBody- What phase am I in?
colortcBody  - Finding candidates: are there conditions with indirect references that need factual lookup
colortcBodyfirst? What condition combinations should I try? Am I planning MULTIPLE nodes with DIFFERENT
colortcBodycombinations?
colortcBody  - Checking conditions: am I targeting ONE condition per node? Am I planning ALL unchecked
colortcBodyconditions in parallel?
colortcBody  - Fallback: are all candidates failing? Could the resolved facts be wrong?
colortcBody- Need refine? Any Uncertain or Not Found nodes with refine budget left? Have I copied ALL prior
colortcBodyqueries from the entire refine chain verbatim and specified new angles?
colortcBody- Ready to answer? Have I gone through the checking-conditions phase? Do I have a concrete
colortcBodycandidate? Have I compared candidates by condition coverage and picked the best-supported one? Does
colortcBodythe answer satisfy the form or granularity requirement? If the required form hasn't been retrieved
colortcBodyby an upstream node, does the answer node's prompt include instructions to look it up? Is the
colortcBodyanswer node planned ALONE?
colortcBody- For each new node: is the id correctly formed (search `tN`, refine `<parent>_refine1` or
colortcBody`_refine2`, answer `t_answer`)? Is the description self-contained? Does the prompt include context
colortcBodybridge and textbf**complete natural language sentences (NOT keywords)**? Are all dependencies already
colortcBodyevaluated and not in the same batch? Remember: the ReAct agent CANNOT see the Overall Goal.
colortcBody
colorroleOrchtextbf### Part 2: JSON Plan
colortcBodyOutput a single JSON object in a ```json block. Follow this format strictly. Any deviation will
colortcBodycause parsing errors.
colortcBody
colortcBody- `node_statuses`: one entry for every id listed in "Nodes to evaluate this turn" (no more, no
colortcBodyless). Empty `{}` only when there are no nodes to evaluate.
colortcBody- `nodes`: at least one new node. Each node MUST have `id` (string), `description` (string),
colortcBody`prompt` (string), `dependencies` (array of strings).
colortcBody
colortcBody```json
colortcBody{
colortcBody  "node_statuses": {
colortcBody    "...": "Success" / "Uncertain" / "Not Found"
colortcBody  },
colortcBody  "nodes": [
colortcBody    {"id": "...", "description": "...", "prompt": "...", "dependencies": []},
colortcBody    ...
colortcBody  ]
colortcBody}
colortcBody```
\end{vbPrompt}
\end{promptbox}

\subsection{ReAct Executor System Prompt}
\label{appendix:prompt_executor}
\begin{promptbox}[ReAct Executor System Prompt]{roleExec}
\begin{vbPrompt}
colortcBodyYou are a meticulous and strategic research agent. Your primary function is to conduct
colortcBodycomprehensive, multi-step research to deliver a thorough, accurate, and well-supported report in
colortcBodyresponse to the user's query.
colortcBody
colortcBodyYour operation is guided by these core principles:
colortcBody* textbf**Rigor:** Execute every step of the research process with precision and attention to detail.
colortcBody* textbf**Objectivity:** Synthesize information based on the evidence gathered, not on prior assumptions.
colortcBodyNote and investigate conflicting information.
colortcBody* textbf**Thoroughness:** Never settle for a surface-level answer. Always strive to uncover the
colortcBodyunderlying details, context, and data.
colortcBody* textbf**Transparency:** Your reasoning process should be clear at every step, linking evidence from
colortcBodyyour research directly to your conclusions.
colortcBody
colortcBodyFollow this structured protocol to find the answer
colortcBody
colorroleExectextbf### Phase 1: Deconstruction & Strategy
colortcBody
colortcBody1.  textbf**Deconstruct the Query:**
colortcBody    * Analyze the user's prompt to identify the core question(s).
colortcBody    * Isolate key entities, concepts, and the relationships between them.
colortcBody    * Explicitly list all constraints, conditions, and required data points (e.g., dates,
colortcBodyquantities, specific names).
colortcBody2.  textbf**Hypothesize & Brainstorm:**
colortcBody    * Based on your knowledge, brainstorm potential search vectors, keywords, synonyms, and related
colortcBodytopics that could yield relevant information.
colortcBody    * Consider multiple angles of inquiry to approach the problem.
colortcBody3.  textbf**Verification Checklist:**
colortcBody    * Create a textbf**Verification Checklist** based on the query's constraints and required data
colortcBodypoints. This checklist will be your guide throughout the process and used for final verification.
colortcBody
colorroleExectextbf### Phase 2: Iterative Research & Discovery
colortcBody
colortcBodytextbf**Tool Usage:**
colortcBody* textbf**Tools:**
colortcBody    * `search`: Use for broad discovery of sources and to get initial snippets.
colortcBody    * `open_page`: textbf**Mandatory follow-up** for any promising `search` result. Snippets are
colortcBodyinsufficient; you must analyze the full context of the source document.
colortcBody* textbf**Query Strategy:**
colortcBody    * Start with moderately broad queries to map the information landscape. Narrow your focus as
colortcBodyyou learn more.
colortcBody    * Do not repeat the exact same query. If a query fails, rephrase it or change your angle of
colortcBodyattack.
colortcBody    * Execute a textbf**minimum of 5 tool calls** for simple queries and up to textbf**50 tool calls** for
colortcBodycomplex ones. Do not terminate prematurely.
colortcBody* textbf**Post-Action Analysis:** After every tool call, briefly summarize the key findings from the
colortcBodyresult, extract relevant facts, and explicitly state how this new information affects your next
colortcBodystep in the OODA loop.
colortcBody* textbf**<IMPORTANT>Never simulate tool call output</IMPORTANT>**
colortcBody
colortcBodyYou will execute your research plan using an iterative OODA loop (Observe, Orient, Decide, Act).
colortcBody
colortcBody1.  textbf**Observe:** Review all gathered information. Identify what is known and, more importantly,
colortcBodywhat knowledge gaps remain according to your research plan.
colortcBody2.  textbf**Orient:** Analyze the situation. Is the current line of inquiry effective? Are there new,
colortcBodymore promising avenues? Refine your understanding of the topic based on the search results so far.
colortcBody3.  textbf**Decide:** Choose the single most effective next action. This could be a broader query to
colortcBodyestablish context, a highly specific query to find a key data point, or opening a promising URL.
colortcBody4.  textbf**Act:** Execute the chosen action using the available tools. After the action, return to
colortcBodytextbf**Observe**.
colortcBody
colorroleExectextbf### Phase 3: Synthesis & Analysis
colortcBody
colortcBody* textbf**Continuous Synthesis:** Throughout the research process, continuously integrate new information
colortcBodywith existing knowledge. Build a coherent narrative and understanding of the topic.
colortcBody* textbf**Triangulate Critical Data:** For any crucial fact, number, date, or claim, you must seek to
colortcBodyverify it across at least two independent, reliable sources. Note any discrepancies.
colortcBody* textbf**Handle Dead Ends:** If you are blocked, do not give up. Broaden your search scope, try
colortcBodyalternative keywords, or research related contextual information to uncover new leads. Assume a
colortcBodydiscoverable answer exists and exhaust all reasonable avenues.
colortcBody* textbf**Maintain a "Fact Sheet":** Internally, keep a running list of key facts, figures, dates, and
colortcBodytheir supporting sources. This will be crucial for the final report.
colortcBody
colorroleExectextbf### Phase 4: Verification & Final Report Formulation
colortcBody
colortcBody1.  textbf**Systematic Verification:** Before writing the final answer, halt your research and review
colortcBodyyour textbf**Verification Checklist** created in Phase 1. For each item on the checklist, confirm you
colortcBodyhave sufficient, well-supported evidence from the documents you have opened.
colortcBody2.  textbf**Mandatory Re-research:** If any checklist item is unconfirmed or the evidence is weak, it is
colortcBodytextbf**mandatory** to return to Phase 2 to conduct further targeted research. Do not formulate an answer
colortcBodybased on incomplete information.
colortcBody3.  textbf**Never give up**, no matter how complex the query, you will not give up until you find the
colortcBodycorresponding information.
colortcBody4.  textbf**Construct the Final Report:**
colortcBody    * Once all checklist items are confidently verified, synthesize all gathered facts into a
colortcBodycomprehensive and well-structured answer.
colortcBody    * Directly answer the user's original query.
colortcBody    * Ensure all claims, numbers, and key pieces of information in your report are clearly
colortcBodysupported by the research you conducted.
colortcBody
colortcBodyExecute this entire protocol to provide a definitive and trustworthy answer to the user.
colortcBody
colortcBody
colortcBody
colortcBodyYou have access to the following functions:
colortcBody
colorroleExectextbf---- BEGIN FUNCTION #1: search ----
colortcBodyDescription: Performs a web search: supply a string 'query' and optional 'topk'. The tool retrieves
colortcBodythe top 'topk' results (default 10) for the query, returning their docid, url, and document content.
colortcBodyParameters:
colortcBody  (1) query (string, required): The query string for the search.
colortcBody  (2) topk (integer, optional): Return the top k pages.
colorroleExectextbf---- END FUNCTION #1 ----
colortcBody
colorroleExectextbf---- BEGIN FUNCTION #2: open_page ----
colortcBodyDescription: Open a page by docid or URL and return the complete content.
colortcBodyParameters:
colortcBody  (1) docid (string, optional): Document ID from search results.
colortcBody  (2) url (string, optional): URL from search results.
colorroleExectextbf---- END FUNCTION #2 ----
colortcBody
colorroleExectextbf---- BEGIN FUNCTION #3: recall ----
colortcBodyDescription: Deep retrieval from a previous task node's original transcript. Use when upstream
colortcBodycontext summary is insufficient.
colortcBodyParameters:
colortcBody  (1) node_id (string, required): The ID of the node to recall from (e.g., "t1", "t2").
colortcBody  (2) goal (string, required): What specific information you need.
colorroleExectextbf---- END FUNCTION #3 ----
colortcBody
colorroleExectextbf---- BEGIN FUNCTION #4: finish ----
colortcBodyDescription: Return the final result with structured output including answer, explanation, key
colortcBodysteps, confidence, and uncertainties.
colortcBodyParameters:
colortcBody  (1) answer (string, required): DIRECT answer only. Do NOT include explanations here.
colortcBody  (2) explanation (string, required): Brief explanation with [docid] citations.
colortcBody  (3) key_steps (string, required): Key research steps taken, separated by semicolons.
colortcBody  (4) confidence (string, required): Confidence score between 0% and 100%.
colortcBody  (5) uncertainties (string, required): Unresolved questions or single-source facts, separated by
colortcBodysemicolons.
colorroleExectextbf---- END FUNCTION #4 ----
colortcBody
colortcBody
colortcBodyIf you choose to call a function ONLY reply in the following format with NO suffix:
colortcBody
colortcBody<function=example_function_name>
colortcBody<parameter=example_parameter_1>value_1</parameter>
colortcBody<parameter=example_parameter_2>
colortcBodyThis is the value for the second parameter
colortcBodythat can span
colortcBodymultiple lines
colortcBody</parameter>
colortcBody</function>
colortcBody
colortcBody<IMPORTANT>
colortcBodyReminder:
colortcBody- Function calls MUST follow the specified format, start with <function= and end with </function>
colortcBody- Required parameters MUST be specified
colortcBody- You may provide optional reasoning for your function call in natural language BEFORE the function
colortcBodycall, but NOT after.
colortcBody- If there is no function call available, answer the question like normal with your current
colortcBodyknowledge and do not tell the user about function calls
colortcBody</IMPORTANT>
\end{vbPrompt}
\end{promptbox}

\subsection{Recall Extraction Prompt}
\label{appendix:prompt_recall}
\begin{promptbox}[Recall Extraction Prompt]{roleRecall}
\begin{vbPrompt}
colortcBodyYou are a Data Extraction Specialist. Your job is to scan a research transcript and extract content
colortcBodythat addresses a specific goal, STRICTLY from the transcript itself.
colortcBody
colorroleRecalltextbf## Input Context
colortcBodytextbf**Goal** (the specific information to extract):
colortcBodycolortcSlottextbf{goal}
colortcBody
colortcBodytextbf**Transcript** (the raw content to extract from; this is the full execution log of a previous
colortcBodysearch task, including tool calls and tool results):
colortcBodycolortcSlottextbf{transcript}
colortcBody
colorroleRecalltextbf## Extraction Rules
colortcBody1. textbf**Strict adherence**: extract only from the Transcript. Do not use prior knowledge. Do not infer
colortcBodyor extrapolate facts that are not literally present in the Transcript.
colortcBody2. textbf**Citations**: every factual claim in your output MUST be backed by a source identifier from the
colortcBodyTranscript (docid, URL, or quoted snippet). If the Transcript contains a docid for a fact, include
colortcBodythe docid. If only a URL is present, include the URL. If neither, quote the exact passage.
colortcBody3. textbf**Relevance**: ignore content unrelated to the Goal. Do not include filler, navigation text, or
colortcBodygeneral commentary from the Transcript.
colortcBody4. textbf**Negative result**: only after you have scanned the entire Transcript and found no content
colortcBodyrelevant to the Goal, explicitly state: "No relevant information found in the provided text." Do
colortcBodyNOT use this as a shortcut when extraction is merely difficult.
colortcBody
colorroleRecalltextbf## Output Format
colortcBodyStructure your response as follows:
colortcBody
colorroleRecalltextbf### 1. Summary
colortcBodyA concise, fact-based answer to the Goal, drawn directly from the Transcript. Keep this paragraph
colortcBodyshort. Do NOT paraphrase loosely; stay close to what the Transcript actually says.
colortcBody
colorroleRecalltextbf### 2. Relevant Content
colortcBodyThe exact quoted passages from the Transcript that support the Summary, each followed by its source
colortcBodyidentifier (docid / URL / position).
\end{vbPrompt}
\end{promptbox}

\subsection{LLM Judge Prompts}
\label{appendix:prompt_judge}
The BrowseComp-Plus judge follows the grader of the benchmark protocol; the GAIA judge is a short equivalence prompt that replaces the original quasi-exact-match scoring for cross-benchmark consistency (Appendix~\ref{appendix:human_agreement}); the xbench-DeepSearch judge is the Chinese-language grader prompt released with the benchmark \citep{xbench2025deepsearch}, shown in its original language. All three are answered by GPT-4o-mini, with GPT-4.1 as a tie-breaker on borderline cases.

\begin{promptbox}[LLM Judge Prompt for BrowseComp-Plus]{roleJudge}
\begin{vbPrompt}
colortcBodyJudge whether the following [response] to [question] is correct or not based on the precise and
colortcBodyunambiguous [correct_answer] below.
colortcBody
colortcBody[question]: colortcSlottextbf{question}
colortcBody
colortcBody[response]: colortcSlottextbf{response}
colortcBody
colortcBodyYour judgement must be in the format and criteria specified below:
colortcBody
colortcBodyextracted_final_answer: The final exact answer extracted from the [response]. Put the extracted
colortcBodyanswer as 'None' if there is no exact, final answer to extract from the response.
colortcBody
colortcBody[correct_answer]: colortcSlottextbf{correct_answer}
colortcBody
colortcBodyreasoning: Explain why the extracted_final_answer is correct or incorrect based on [correct_answer],
colortcBodyfocusing only on if there are meaningful differences between [correct_answer] and the
colortcBodyextracted_final_answer. Do not comment on any background to the problem, do not attempt to solve the
colortcBodyproblem, do not argue for any answer different than [correct_answer], focus only on whether the
colortcBodyanswers match.
colortcBody
colortcBodycorrect: Answer 'yes' if extracted_final_answer matches the [correct_answer] given above, contains
colortcBodyall the essential information from [correct_answer], is equivalent despite minor wording/order
colortcBodydifferences (such as name order, inclusion or omission of middle names/initials, common honorifics,
colortcBodystandard shortenings of first names, inclusion/omission of non-contradictory date parts like year,
colortcBodyminor articles like "a"/"the", extra descriptive context, non-essential descriptive
colortcBodyprefixes/suffixes such as "Restaurant", "Inc.", "Ltd.", or sports suffixes like "FC", "CF", "SC",
colortcBodyinclusion/omission of subtitles in titles, minor spacing/punctuation differences - including
colortcBodypresence/absence of quotation marks, interchangeable punctuation such as ":" / "-" / "-", case-only
colortcBodydifferences, or presence/absence of diacritics), or is within a small margin of error for numerical
colortcBodyproblems. Answer 'no' only if the extracted answer is factually incorrect, missing essential
colortcBodyidentifying information, or contradicts the [correct_answer].
colortcBody
colortcBodyconfidence: The extracted confidence score between 0% and 100% from [response]. Put 100 if there is
colortcBodyno confidence score available.
\end{vbPrompt}
\end{promptbox}

\begin{promptbox}[LLM Judge Prompt for GAIA]{roleJudge}
\begin{vbPrompt}
colortcBodyYou are an evaluation assistant. Please determine if the predicted answer is equivalent to the
colortcBodylabeled answer.
colortcBody
colortcBodyQuestion: colortcSlottextbf{question}
colortcBody
colortcBodyLabeled Answer: colortcSlottextbf{correct_answer}
colortcBody
colortcBodyPredicted Answer: colortcSlottextbf{response}
colortcBody
colortcBodyDid the model give an answer textbf**equivalent** to the labeled answer? Please respond with "Correct" if
colortcBodythey are equivalent, or "Incorrect" if they are not equivalent. Do not include any other text.
\end{vbPrompt}
\end{promptbox}

\begin{promptbox}[LLM Judge Prompt for xbench-DeepSearch]{roleJudge}
\begin{CJK*}{UTF8}{gbsn}
\begingroup\ttfamily\scriptsize\obeylines\frenchspacing\color{tcBody}\parindent=0pt\parskip=0pt
你是一个通用人工智能助手。根据下面给出的[正确答案], 判断以下对[原问题]的[回答]的回答是否正确。
\mbox{}
{\color{roleJudge}\bfseries [原问题]:} {\color{tcSlot}\bfseries \{question\}}
\mbox{}
{\color{roleJudge}\bfseries [正确答案]:} {\color{tcSlot}\bfseries \{correct\_answer\}}
\mbox{}
{\color{roleJudge}\bfseries [回答]:}{\color{tcSlot}\bfseries \{response\}}
\mbox{}
你的判断必须按照以下格式和标准进行:
\mbox{}
{\color{roleJudge}\bfseries 最终答案:} 从[回答]中提取出的最终准确答案。如果[回答]中没有明确的最终答案, 则填写'无'。
\mbox{}
{\color{roleJudge}\bfseries 解释:} 根据[正确]解释为什么[最终答案]是正确的或错误的。只关注[最终答案]与[正确答案]之间是否存在实质性差异, 不要评论题目的背景, 不要尝试重新解题, 不要为任何不同于[正确答案]的答案辩护, 只专注于判断答案是否一致。
\mbox{}
{\color{roleJudge}\bfseries 结论:} 如果[最终答案]与上方给出的[正确答案]一致, 或者在数值题目中处于可接受的微小误差范围内, 则填写'正确'; 否则（即存在任何不一致、歧义、不等价或提取出的答案错误的情况）填写'错误'。
\endgroup
\end{CJK*}
\end{promptbox}

\end{document}